\documentclass[lettersize,journal]{IEEEtran}
\usepackage{amsmath,amsfonts}
\usepackage[ruled]{algorithm2e}
\usepackage{array}
\usepackage{textcomp}
\usepackage{stfloats}
\usepackage{url}
\usepackage{verbatim}
\usepackage{graphicx}
\usepackage{cite}
\usepackage{multirow}
\usepackage{arydshln}
\usepackage{bm}
\usepackage{bbding}
\usepackage{dashrule}
\usepackage{booktabs}
\usepackage{float}  %设置图片浮动位置的宏包
\usepackage{dsfont}
\usepackage{enumerate}
\usepackage{microtype}
\usepackage{amssymb}
\usepackage{makecell}
\usepackage{color}
\usepackage{lineno,hyperref}
\usepackage{tablefootnote}

\begin{document}

\title{DATA: Multi-Disentanglement based Contrastive Learning for Open-World Semi-Supervised Deepfake Attribution}

\author{Ming-Hui Liu,~Xiao-Qian Liu,~Xin Luo,~Xin-Shun Xu*,~\IEEEmembership{Senior Member,~IEEE}
        % <-this % stops a space
\thanks{This work was supported in part by: (1) National Natural Science Foundation of China (Grant No. 62172256, No. 62202278, and No. 62202272); (2)
Natural Science Foundation of Shandong Province (Grant No. ZR2019ZD06). (Corresponding author: Xin-Shun Xu.)

Ming-Hui Liu, Xiao-Qian Liu, and Xin Luo are with the School of Software, Shandong University, Jinan 250101, China (e-mail: liuminghui@mail.sdu.edu.cn; jlrxqxq370322@126.com; luoxin.lxin@gmail.com).
Xin-Shun Xu is with the School of Software, Shandong University, Jinan 250100, China, and also with the Quan Cheng Laboratory, Jinan 28666, China (e-mail: xuxinshun@sdu.edu.cn)}% <-this % stops a space
\thanks{Manuscript received July 11, 2024}}

\markboth{Journal of \LaTeX\ Class Files,~Vol.~14, No.~8, August~2021}%
{Shell \MakeLowercase{\textit{et al.}}: A Sample Article Using IEEEtran.cls for IEEE Journals}


\maketitle

\begin{abstract}
Deepfake attribution (DFA) aims to perform multiclassification on different facial manipulation techniques, thereby mitigating the detrimental effects of forgery content on the social order and personal reputations. However, previous methods focus only on method-specific clues, which easily lead to overfitting, while overlooking the crucial role of common forgery features. Additionally, they struggle to distinguish between uncertain novel classes in more practical open-world scenarios. To address these issues, in this paper we propose an innovative multi-DisentAnglement based conTrastive leArning framework, DATA, to enhance the generalization ability on novel classes for the open-world semi-supervised deepfake attribution (OSS-DFA) task. Specifically, since all generation techniques can be abstracted into a similar architecture, DATA defines the concept of `Orthonormal Deepfake Basis' for the first time and utilizes it to disentangle method-specific features, thereby reducing the overfitting on forgery-irrelevant information. Furthermore, an augmented-memory mechanism is designed to assist in novel class discovery and contrastive learning, which aims to obtain clear class boundaries for the novel classes through instance-level disentanglements. Additionally, to enhance the standardization and discrimination of features, DATA uses bases contrastive loss and center contrastive loss as auxiliaries for the aforementioned modules. Extensive experimental evaluations show that DATA achieves state-of-the-art performance on the OSS-DFA benchmark, e.g., there are notable accuracy improvements in $2.55\% / 5.7\%$ under different settings, compared with the existing methods.
\end{abstract}


\section{Introduction}
\begin{figure}[t]
    \centering
    \includegraphics[width=8.6cm]{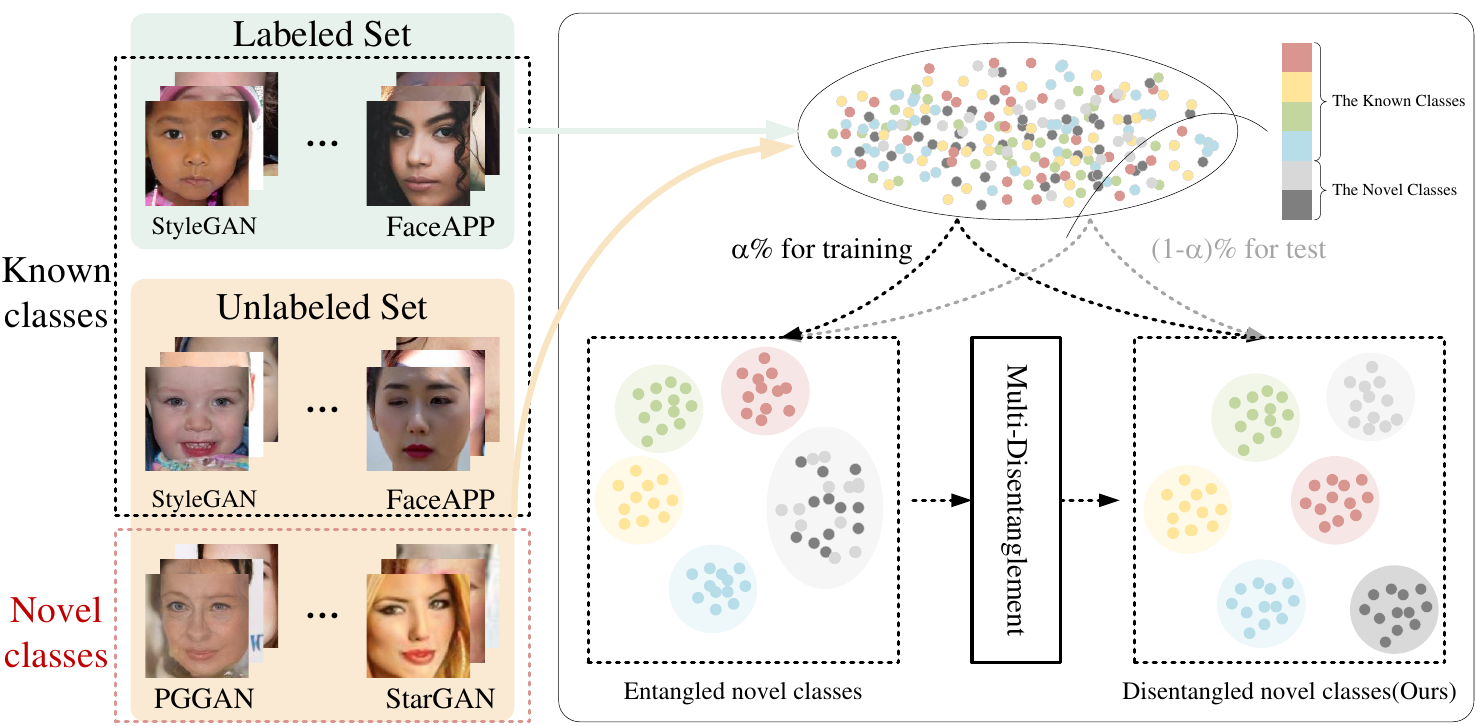}
    \caption{The OSS-DFA benchmark consists of diverse known and novel deepfake classes. Meanwhile, the training set and test set share the same label space and our goal is to disentangle interwoven novel samples.}\label{fig.1}
\end{figure}

\IEEEPARstart{T}{he} swiftly advancing deepfake technologies \cite{8,12}, which deceive the public by manipulating facial videos, have severely undermined social trust mechanisms. To cope with such security threats, many deepfake detection (DFD) works are devoted to simply determining whether a given face is fake or real \cite{17,20,21,22,ojha2023towards,doloriel2024frequency,chai2020makes} by establishing unified prototypes. However, the impacts of different techniques are so distinct that focusing solely on binary classification has lots of limitations for downstream tasks. For example, simple attribute manipulation can only change the makeup of characters, whereas a completely inauthentic face-swap video will spawn rumors and result in devastating consequences. Therefore, Deepfake Attribution (DFA), a novel multi-classification task, has emerged. It tries to further categorize different deepfake techniques relying on method-specific information. Moreover, in contrast to just informing users of a general result, showing an exact deepfake attribution result will be more convincing. 
Unfortunately, obtaining accurate annotations for proliferate deepfake technologies is labor intensive, and this annotation burden has driven the community to make full use of unlabeled samples, which come from not only the known (labeled) classes but also the unseen novel classes. To meet the above practical needs, Open-World Semi-Supervised Deepfake Attribution (OSS-DFA), which discovers novel classes and allocates correlative samples without any additional annotation workload, has become a new tendency. As illustrated in Fig.~\ref{fig.1}, both labeled and unlabeled data are utilized during training, and the test set shares the same label space with the training set \cite{1}.

Although traditional open-world semi-supervised learning methods \cite{25-1,25-2} have achieved high accuracy on novel classes by leveraging diverse prior knowledge across various application scenarios, they still cannot cope with the complex OSS-DFA task, in which forgery-irrelevant pattern information tends to play an undesirable dominant role during classification. This is because previous open-world methods assume that different classes share the same distribution, but the OSS-DFA task always suffers from an inherent data issue: different deepfake classes exhibit significant forgery-irrelevant data biases (diverse resolutions, semantic contents, and viewpoints) in limited training set \cite{28}. Therefore, focusing on the differences between deepfake techniques makes OSS-DFA models even more sensitive to these biases, and prone to secretly relying on them. This means that even if a model can obtain high accuracy through overfitting behavior during the training phase, it may become invalid when tested on novel classes due to its poor generalization. Although performing a data-level enhancement can significantly enrich data diversity and improve generalization ability \cite{29,30}, it still has performance bottlenecks due to finite data processing techniques and computational capability. Other works have attempted to avoid overfitting with the help of local or high-frequency information, but ensuring accurate attention to forgery-related information is still difficult. Even worse, almost all the DFA methods are unable to handle novel classes in open-world scenarios or need a predefined number for novel classes.

To fundamentally overcome the obstacle to enhancing generalization, we rethink the architectures of common generator models \cite{13,16,8-2}, and find that they can be abstracted into the same paradigm. Overall, to achieve realistic performance, the generator network first maps the input data into a high dimension. Subsequently, a decoder module produces the generation via an up-sampling operation, and then a series of concatenation and blur operations merge the generation with original background information as the final output. Building upon the above insight, we introduce the concept of `Orthonormal Deepfake Basis' for the first time and suppose that all of the deepfake artifacts can be formed by the same set of bases. This means that the network can be forced to focus on forgery artifacts, rather than overfitting to forgery-irrelevant data biases by exploiting the method-general deepfake bases. 

To address the challenges in the OSS-DFA task, we propose a novel Multi-Disentanglement based Contrastive Learning framework, DATA, to disentangle method-specific features from the feature level, and disentangle interwoven unlabeled samples from the instance level: 
1)~We introduce the Deepfake Basis Exploration~(DBE) module to force the focus on forgery-related features by exploiting method-general deepfake bases. In this module, a set of `Orthonormal Deepfake Bases' are produced by the Gram-Schmidt Process and then method-specific features can be disentangled with the collaboration of deepfake bases and related Bases Contrastive Loss $\mathcal{L}_{Bases}$.
2)~To further disentangle unlabeled samples, we leverage an Augmented-Memory based Clustering~(AMC) module from a more macroscopic perspective. Recognizing that pre-defining the number of novel classes is impractical, we construct an Augmented-Memory mechanism to gradually discover and store novel class prototypes. Meanwhile, with reference to the memory, the noise in pseudo-labels is significantly alleviated. Moreover, we enhance the discrimination of samples by gathering them to stored prototypes using Center Contrastive Loss $\mathcal{L}_{Center}$ based on the anchor nature. At this point, the unlabeled samples are disentangled into the appropriate classes.

Overall, our contributions can be summarized as follows:
\begin{itemize}
\item We introduce a novel framework, DATA, for Open-World Semi-Supervised Deepfake Attribution (OSS-DFA) task, which enhances the performance through both feature-level and instance-level disentanglement. 

\item We rethink the common generator architectures and define the concept of `Orthonormal Deepfake Basis' for the first time. Based on the constraint of deepfake bases in features excavation, we further construct a Deepfake Basis Exploration~(DBE) module to disentangle the desirable method-specific forgery-related features. 

\item We design an Augmented-Memory based Clustering~(AMC) module, in which we gradually discover and store novel class prototypes via an Augmented-Memory mechanism and disentangle unlabeled samples into appropriate classes by ingenious contrastive learning. 

\item Extensive experiments are conducted. The results show that our method has a large advantage over the existing OSS-DFA methods.  In particular, for novel classes, it achieves a state-of-the-art performance across different experimental settings.
\end{itemize}

\section{Related work}

\begin{figure*}[t]
    \centering
    \includegraphics[width=1\textwidth]{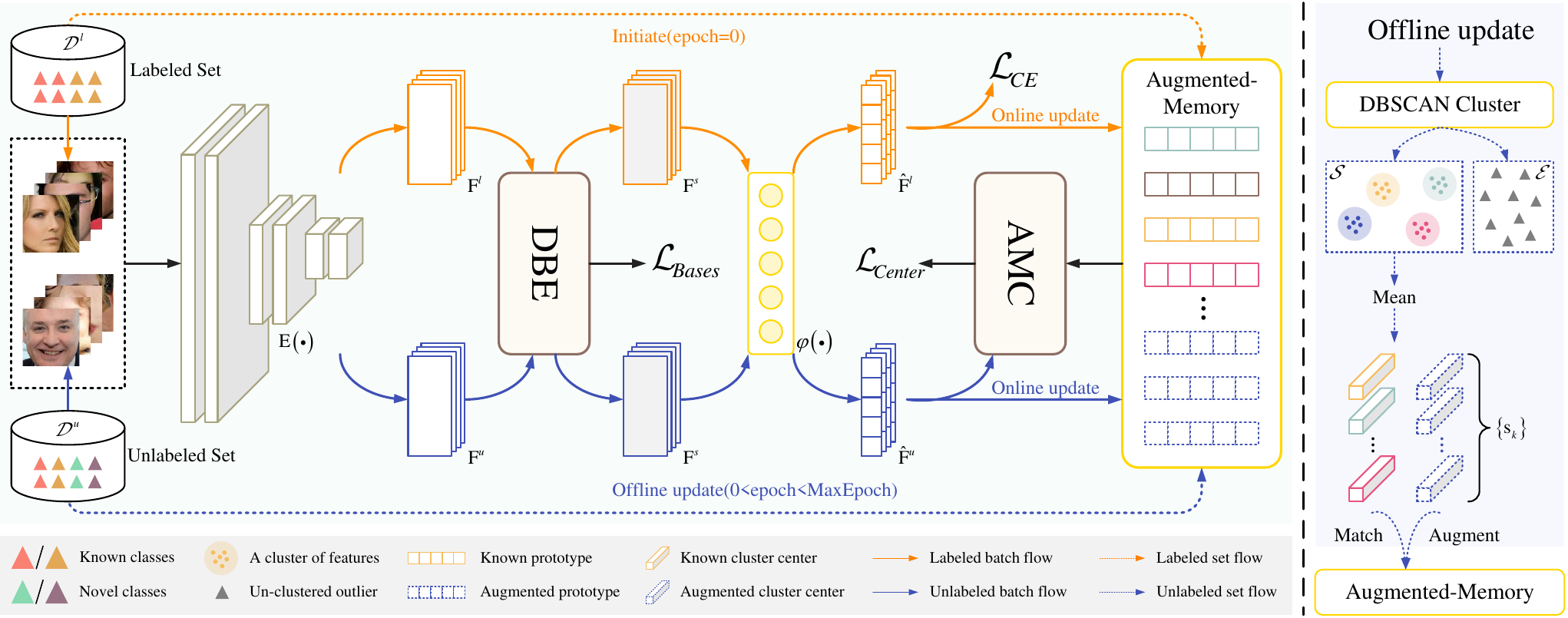}
    \caption{An overview of the DATA framework. The DBE module is used to explore the `orthonormal deepfake basis set' and disentangle method-specific feature $\mathbf{f}^s$. The AMC module is used to conduct the instance-level disentanglement with the assistance of an Augmented-Memory.}\label{fig:DATE}
\end{figure*}

\subsection{Deepfake Detection and Attribution}
To reduce the annotation cost, deepfake detection (DFD) commonly attempts to obtain a universal model and generalizes it to different target domains during the test stage. Hence, some works try to directly extract temporal or frequency information~\cite{61,63} which is nearly impacted by domain discrepancies. For instance, Haliassos et al.~\cite{2} incorporated the time dimension by acquiring time-dense features. Wang et al.~\cite{3} divided network weights into spatial group or temporal group, and alternately froze them to focus on diverse features. Wang et al.~\cite{4} exploited the relation inference capability of graph convolution network to unearth more generalizable relationships. Another type of work aims to avoid overfitting to biased training data~\cite{5,64}. Ba et al.~\cite{6} employed mutual information theory and extracted multiple non-overlapping local features. Huang et al.~\cite{7} differentiated facial information into implicit and explicit identities to learn deepfake essence. Unfortunately, the aforementioned binary classification methods cannot be directly applied to multi-classification deepfake attribution (DFA) tasks and almost all DFA methods still overlook the generalization issue. Yu et al.~\cite{7-1} relied on pre-assigned fingerprints to attribute the generative models. Yang et al.~\cite{7-2} can only attribute samples to coarse source architectures. Yu et al.~\cite{7-4} defined the model fingerprint per GAN instance. Sun et al.~\cite{1} introduced the open-world semi-supervised deepfake attribution (OSS-DFA) benchmark but they overlook the dataset bias and rely on presuppositions. In contrast, our method balances the accuracy and generalization during training by excavating method-general orthogonal deepfake bases and designing an applicable Augmented-Memory module for uncertain novel classes.

\subsection{Open-world Semi-Supervised Learning}
Unlike closed-set semi-supervised learning (SSL), open-world semi-supervised learning (OW-SSL) utilizes both labeled and unlabeled samples for training while the unlabeled samples may originate not only from known classes but also from previously unseen (novel) classes. With the increasing data demand, OW-SSL has become an excellent alternative to alleviate the annotation burden and numerous relevant methods have emerged \cite{27-1}. Sun et al.~\cite{25-4} formulated the OW-SSL task as a clustering problem and attempted to reduce the error rate by a spectral approach. Wang et al.~\cite{25-5} employed multi-granularity semantic concepts to mitigate the noise of pseudo-labels. Rizve et al.~\cite{25-6} transformed the OW-SSL task into a standard SSL task by novel class discovery. Despite the promise, the above OW-SSL methods almost overlooked the data bias in known classes which is especially intractable in OSS-DFA tasks. Xiao et al.\cite{25-3} realized the data issue and utilized the neural collapse phenomenon to calibrate the distribution of novel classes, but they still assume the number of novel classes as a known apriori concept. In contrast, our DATA framework does not require additional prior knowledge and mitigates known class biases by focusing on forgery-general `Orthonormal Deepfake Bases'.

\subsection{Dataset Bias}
As data become increasingly important, dataset bias has also caught much attention from the community in recent years. The methods for dataset bias elimination can be divided into two major categories: a direct way of refining training data \cite{53} and an indirect way of enhancing the adaptability of models to biased data \cite{55}. For example, Torralba et al.~\cite{52} analyzed and compared the status of existing datasets and identified an improved way of dataset collection as well as evaluation protocols. Clark et al.~\cite{57} treated dataset bias as prior knowledge and used it to encourage the focus on other patterns in the data, thereby enhancing the generalization of the network. Bras et al.~\cite{59} proposed an adversarial approach to filter dataset bias during feature extraction, effectively addressing the prevalent overestimation. Khosla et al.~\cite{60} simultaneously learned two sets of weights and utilized the visual world weights to approximate the ideal weights obtained from an unbiased dataset. Although debiasing has made significant contributions in various fields, it has not been explicitly considered in deepfake attribution, where the dataset is combined by multiple sub-datasets. Thus, if no additional measures are taken, models will combine the differences in deepfake techniques with dataset bias for final classification and significantly damage the generalization performance. For this reason, we propose a novel framework based on the concept of `Orthonormal Deepfake Basis', and force the network to extract only unbiased forgery-relevant features for the OSS-DFA task.

\section{METHODOLOGY}
\subsection{Overview}
In order to accomplish Open-World Semi-Supervised Deepfake Attribution (OSS-DFA), we propose a Multi-Disentanglement based Contrastive Learning (DATA) method. The overall pipeline of our proposed framework is illustrated in Fig.\ref{fig:DATE}. During training, we take the labeled dataset ${\cal{D}}^l = {\{x_i,y_i\}}_{i=1}^N$ containing known classes, and the unlabeled dataset ${\cal{D}}^u = \{x_i\}_{i = 1}^M$ containing all classes, as input. The classes in labeled and unlabeled sets are ${\cal{C}}_{known}=\{1,2,\cdots,l\}$ and ${\cal{C}}_{all}=\{1,2,\cdots,u\}$ (${\cal{C}}_{known} \subset {\cal{C}}_{all}$). We also define the novel classes as ${\cal{C}}_{novel} = {\cal{C}}_{all}\setminus{\cal{C}}_{known}$.
Overall, after basic feature extraction, we utilize the Deepfake Basis Exploration~(DBE) module to disentangle the method-specific and forgery-related feature ${\mathbf f}^s$ at the feature level for deepfake attribution. Then, in subsequent Augmented-Memory based Clustering~(AMC) module, we disentangle unlabeled samples at the instance level by continuously generating novel prototypes. In addition, we also set the corresponding Bases Contrastive Loss $\mathcal{L}_{Bases}$ and Center Contrastive loss $\mathcal{L}_{Center}$ to assist with optimization. With the above endeavors, we enhance the discrimination accuracy between the novel deepfake classes.

\subsection{Deepfake Basis Exploration}
\begin{figure}[t]
    \centering
    \includegraphics[width=0.48\textwidth]{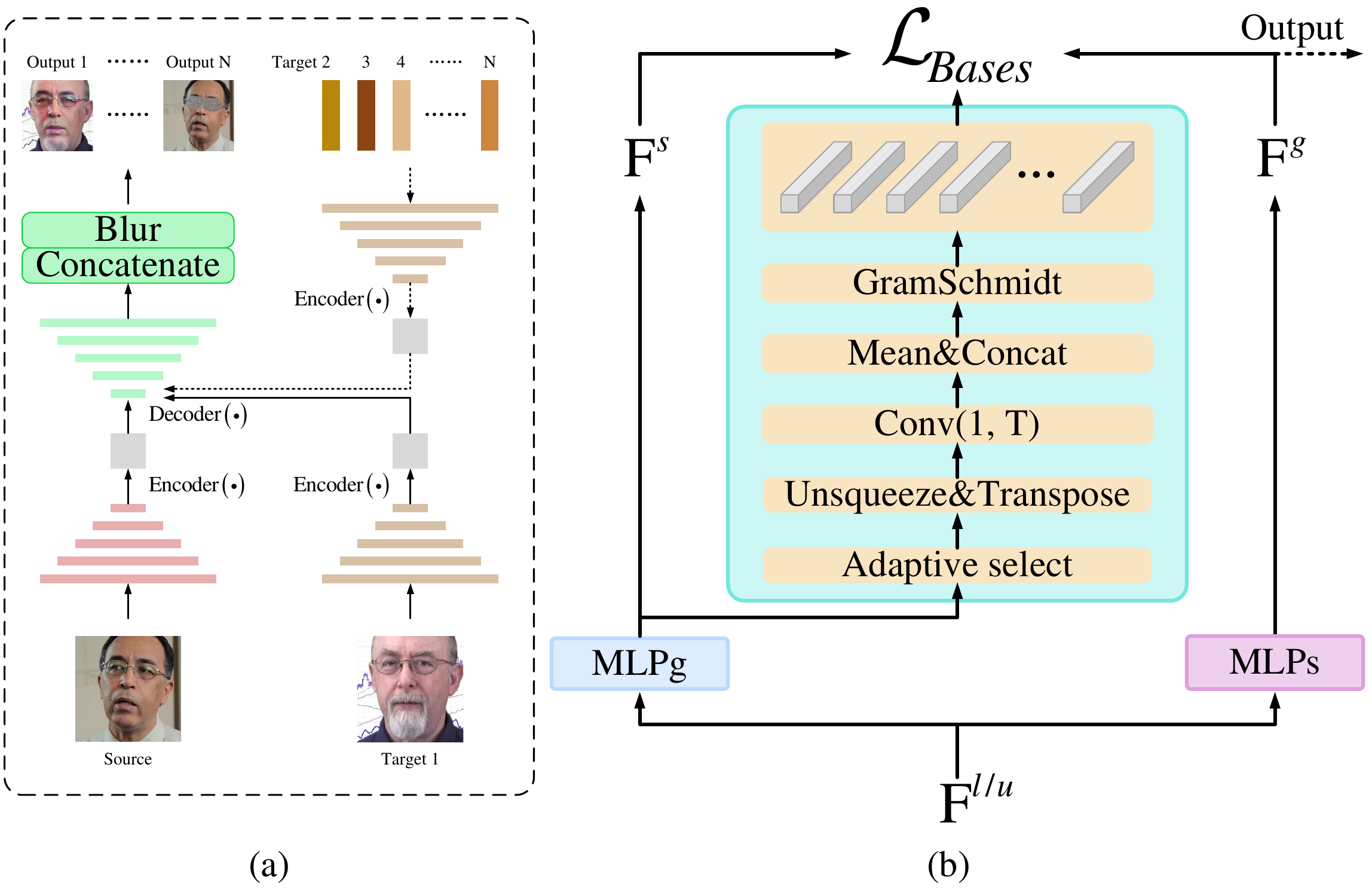}
    \caption{An overall pipeline of the DBE module.}\label{fig:DBE}
\end{figure}

To alleviate the inherent dataset bias between fake classes, we deviate from the limited data augmentation strategy. Instead, we find that almost generator models share a similar architecture, as shown in Fig.~\ref{fig:DBE}(a) and the generation process can be summarized as follows: extracting source features; upsampling features to target dimensions and concatenating them with original images; Blurring the concatenation edges, and obtaining the final outputs. Driven by this, we propose a novel hypothesis: each deepfake technique can be represented by the same method-general `orthogonal deepfake basis set'. Moreover, owing to the natural relationship, explicit exploration of the `orthogonal deepfake basis set' reversely enhances the focus on forgery-related features. Based on the above assumption, we construct a Deepfake Basis Exploration~(DBE) module to disentangle unbiased method-specific features for OSS-DFA tasks, which incorporate more truly forgery-relevant features rather than superficial dataset biases. Firstly, we integrate frame sampling with face localization for facial frames, and perform feature extraction in a mini-batch to obtain a labeled feature set ${\mathbf{F}^l}=\{\mathbf{f}_1^l, \mathbf{f}_2^l, \cdots, \mathbf{f}_n^l\}$ and an unlabeled feature set ${{\mathbf{F}^u}=\{\mathbf{f}_1^u,\mathbf{f}_2^u, \cdots, \mathbf{f}_m^u\}}$. Then, we employ the GramSchmidit process to learn an `orthogonal deepfake basis set' and use it to refine the original features under the supervision of $\mathcal{L}_{Bases}$. The above process is illustrated in Fig.~\ref{fig:DBE}(b).

\textbf{Definition (Gram–Schmidt Process).} In mathematics, if a set of vectors in an inner product space can form a subspace, these vectors are termed the bases of the subspace. The Gram-Schmidt process provides a method to derive the orthogonal basis set. Given $k$ vectors $\{{\mathbf v}_1, {\mathbf v}_2, \cdots, {\mathbf v}_k\}$, the Gram–Schmidt process defines the bases $\{{\mathbf u}_1, {\mathbf u}_2, \cdots, {\mathbf u}_k\}$ as follows:
\begin{equation}
\begin{split}
&{{{\mathbf u}_1} = {{\mathbf v}_1}}\\
&{{{\mathbf u}_2} = {{\mathbf v}_2} - {{\mathbf {proj}}_{{\mathbf u}_1}}({\mathbf v}_2)}\\
&{{{\mathbf u}_3} = {{\mathbf v}_3} - {{\mathbf {proj}}_{{\mathbf u}_1}}({\mathbf v}_3) - {{\mathbf {proj}}_{{\mathbf u}_2}}({\mathbf v}_3)}\\
&\cdots\\
&{{{\mathbf u}_k} = {{\mathbf v}_k} - \sum\limits_{j = 1}^{k - 1} {{{\mathbf {proj}}_{{\mathbf u}_j}}({\mathbf v}_k)}}
\end{split},
\end{equation}
where $\mathbf {proj_u}(\mathbf v)$ denotes the vector projection of a vector $\mathbf v$ on a nonzero vector $\mathbf u$.

\textbf{Learning orthogonal deepfake bases by Gram–Schmidt.} For the purpose of deepfake attribution, we naturally aim to disentangle the distinctive method-specific features. To achieve this goal, we employ a general-MLP module $\mathrm{MLP}_{g}(\cdot)$ and specific-MLP module $\mathrm{MLP}_{s}(\cdot)$ \cite{6}, which have the same structure but different parameters, to differentiate the method-specific feature $\mathbf{f}^s$ and method-general feature $\mathbf{f}^g$ from the original feature $\mathbf{f}$:
\begin{equation}
\begin{split}
&\mathbf{f}^g = \mathrm{MLP}_{g}(\mathbf{f}) \\
&\mathbf{f}^s = \mathrm{MLP}_{s}(\mathbf{f}). \\
\end{split}
\end{equation}
Here, for brevity, we omit the superscripts of $\mathbf{f}^l$/$\mathbf{f}^u$ and represent both of them by $\mathbf{f}$. 

However, due to the absence of supervision, $\mathbf{f}^s$ and $\mathbf{f}^g$ suffer from collapsing to the same representation. To remedy this, we further refine them by resorting to learning a set of universal orthogonal deepfake bases. First, we average $n'$ labeled method-general deepfake features $\mathbf{F}^g=\{ \mathbf{f}_1^g, \mathbf{f}_2^g, \cdots,\mathbf{f}_{n'}^g\}(n' \le n)$ in a mini-batch to obtain a global deepfake feature $\mathbf{\bar f}^g$. Subsequently, we decompose $\mathbf{\bar f}^g$ along the channel dimension and yield $T$ feature components. Finally, we utilize Gram–Schmidt \cite{35} process to refine the aforementioned components and obtain an orthogonal deepfake basis set ${\mathbf{O}}=\{\mathbf{o}_1,\mathbf{o}_2, \cdots,\mathbf{o}_T\}$:
\begin{equation}
\begin{split}
  & \mathbf{\bar f}^g = \phi ({\mathrm{Mean}}(\mathbf{F}^g)) \\
  & {\mathbf{O}} = {\mathrm{GramSchmidt}}({\mathrm{Conv}}(\mathbf{\bar f}^g)). \\
\end{split}
\end{equation}
Here, ${\mathrm{Mean}}(\cdot)$ takes an average across batch-size dimension; $\phi(\cdot)$ is a dimension transformation operation; $\mathrm{Conv}(\cdot)$ is a single convolutional layer with an input channel of $1$ and an output channel of $T$. For computational simplicity, we concatenate and integrate all of the orthogonal deepfake bases using $\mathrm{Concat}(\cdot)$ along the channel dimension to obtain an aggregation representation $\tilde {\mathbf{o}}$ with the same dimensions as $\mathbf{f}:$
\begin{equation}
\begin{split}
  & \tilde {\mathbf{o}} = \mathrm{Concat}({\mathbf{O}}). \\
\end{split}
\end{equation}
Obviously, by exploiting such deepfake bases, $\mathrm{MLP}_{g}(\cdot)$ output more method-general and highly forgery-related information that is different from the output of $\mathrm{MLP}_{s}(\cdot)$.

\textbf{Bases Contrastive Loss.} 
To disentangle accurate method-specific features for the OSS-DFA task, we introduce a Bases Contrastive Loss to achieve feature-level contrastive learning. For this purpose, we move $\mathbf{f}^g$ to deepfake bases $\tilde {\mathbf{o}}$ while move away from $\mathbf{f}^s$ and above process can be achieve as follows:
\begin{equation}
\label{eq:loss_base}
{\mathcal L}_{Bases} = \frac{1}{m+n} \sum \limits_{i=0}^{m+n} 
 \max(0, \tau + {\|\mathbf{f}^g-\tilde {\mathbf{o}}\|}_2 - 
{\|\mathbf{f}^g - \mathbf{f}^s\|}_2),
\end{equation}
where $m+n$ is the total number of samples in a mini-batch, and $\tau$ is a boundary hyperparameter that controls the sensitivity of the loss function. At this point, method-general features and method-specific features are gradually separated into $\mathbf{f}^g$ and $\mathbf{f}^s$. Meanwhile, because using the same source, $\mathbf{f}^s$ also contains more unbiased forgery information under the guidance of deepfake highly relevant feature $\mathbf{f}^g$. Finally, DBE module integrates the fully connected layer $\mathrm{FC(\cdot)}$ and normalization layer $\mathrm{Softmax(\cdot)}$ to form the deepfake attribution classifier $\varphi(\cdot)=\mathrm{Softmax(\mathrm{FC(\cdot)})}$ and obtain the corresponding classification probability $\mathbf{\hat f}$: 
\begin{equation}
{\mathbf{\hat f}}=\varphi({\mathbf{f}}^s).
\end{equation}
Then, the subsequent module will only utilize ${\mathbf{\hat f}}$ for novel class discovery and semi-supervised contrastive learning.

\subsection{Augmented-Memory based Clustering}
\begin{figure}[t]
    \centering
    \includegraphics[width=0.48\textwidth]{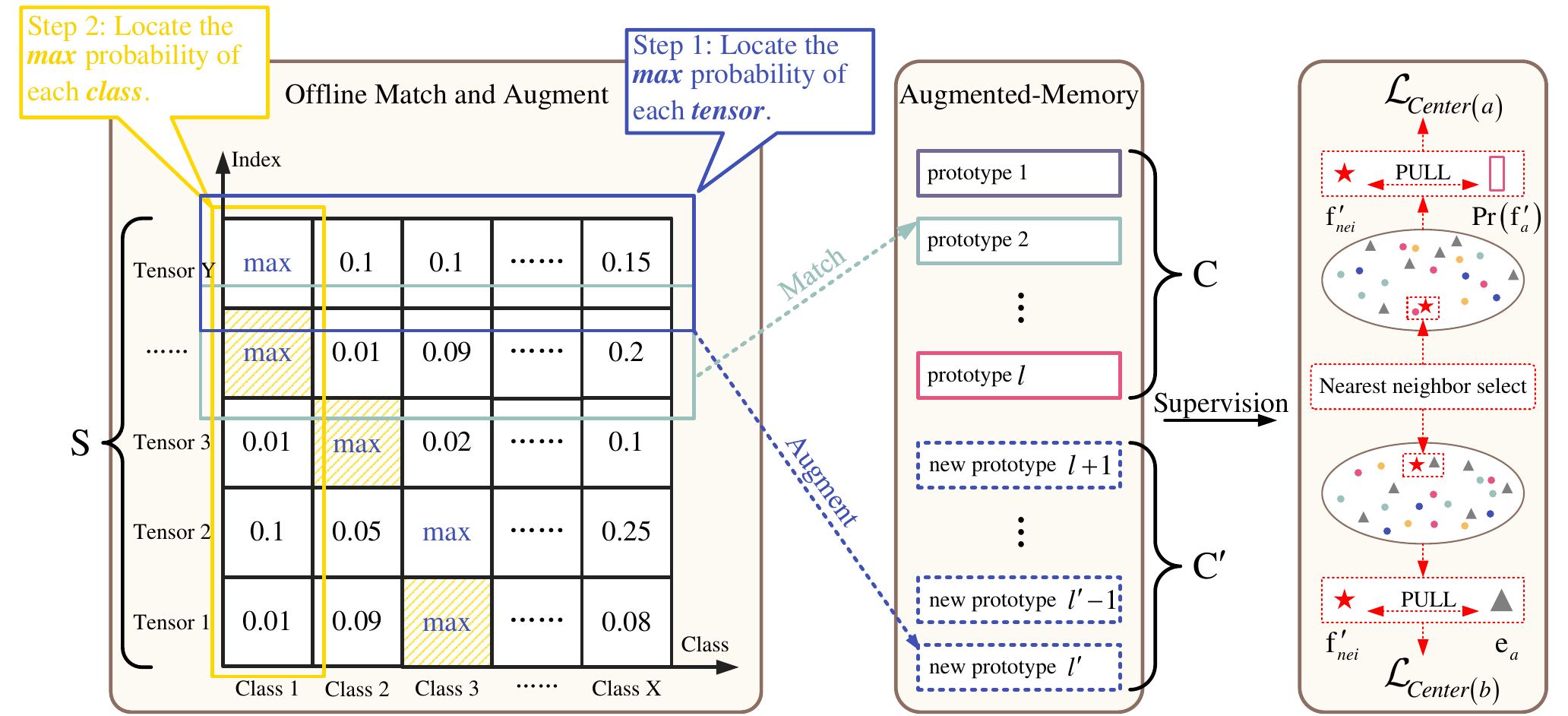}
    \caption{A brief schematic of the AMC module.}\label{fig:AMCN}
\end{figure}
Owing to the absence of novel class supervision, the learned features tend to overfit to known classes, and the unlabeled samples will be intertwined. To alleviate this and further utilize the unlabeled samples, we construct an Augmented-Memory based Clustering (AMC) module for novel class discovery and pseudo-label generation. Then, through semi-supervised contrastive learning, the remaining outliers are assigned to corresponding classes from easy to hard as an instance-level disentanglement process.

\textbf{DBSCAN Cluster.} 
To utilize the underlying information of unlabeled features, we conduct DBSCAN clustering \cite{37-2} on all of the unlabeled features $\mathbf{\hat F}^u = \{ {\mathbf{\hat f}_u^l,\mathbf{\hat f}_2^u, \cdots ,\mathbf{\hat f}_N^u} \}$ among the whole training set before each epoch. Thereafter, based on the clustering results, unlabeled features can be divided into a random number of un-clustered outlier instances ${\cal E}=\{\mathbf{e}_1,\mathbf{e}_2, \cdots \}$ and clusters ${\cal S}=\{S_1,S_2,\cdots \}$, where each $S_k$ contains a number of similar features:
\begin{equation} 
{\cal E}, {\cal S}=\mathrm{DBSCAN}(\mathbf{\hat F}^u).
\end{equation}
Unlike the benchmark, we opt for the more flexible DBSCAN clustering instead of the K-means clustering, which requires the number of novel classes ${\cal C}_{novel}$ as known priori. Obviously, presupposing the above knowledge is unrealistic under an open-world scenario. After the clustering process, we assign the same pseudo label to samples within the same cluster. Then, we compute the cluster centroid of cluster $S_k$ and denoted the centroid feature as ${\mathbf s}_k$:
\begin{equation}
{\mathbf s}_k = \frac{1}{|S_k|}\sum\limits_{\mathbf{\hat f}_i^u \in {S_k}}
\mathbf{\hat f}_i^u.
\end{equation}
Here, $k$ is the cluster index; $|\cdot|$ denotes the number of features in a cluster.

\textbf{Memory Initialization.} 
To address severe overfitting issues during training while remedying the detrimental impact of pseudo-label noise \cite{38}, we utilize a memory mechanism \cite{36} to preserve the known class prototypes, which serve as an important reference for discovering novel classes. To accomplish this, before the training stage, we average all of the labeled features $\mathbf{\hat F}^l = \{ {\mathbf{\hat f}_1^l,\mathbf{\hat f}_2^l, \cdots ,\mathbf{\hat f}_N^l} \}$ in training set according to their categories to obtain a set of class prototypes ${\mathbf C}=\{{\mathbf c}_1,{\mathbf c}_2, \cdots, {\mathbf c}_l\}$ as the initial content of memory:
\begin{equation}
{{\mathbf c}_k} = \frac{1}{|\mathbf{\hat F}_k^l|}\sum\limits_{\mathbf{\hat f}_i^l \in \mathbf{\hat F}_k^l} \mathbf{\hat f}_i^l,
\end{equation}
where $k$ is the category index; set $\mathbf{\hat F}_k^l$ contains all labeled features belonging to category $k$ ($\mathbf{\hat F}_k^l \subset \mathbf{\hat F}^l$ and $\mathbf{\hat F}_k^l \ne \varnothing$); $|\cdot|$ denotes the number of features in a particular set. 

\textbf{Novel Class Discovery.} 
Considering that the number of novel classes in the OSS-DFA task is uncertain, we design a more suitable Augmented-Memory module to store a series of known and novel prototypes, the number of which gradually increases along with training. The process is described in Fig.\ref{fig:AMCN}. 

Before every epoch, to discover and append novel prototypes for Augmented-Memory accurately, we utilize clustering centers ${\mathbf{S}}=\{{\mathbf s}_1, {\mathbf s}_2,\cdots\}$ as candidates and attempt to discover reliable novel prototypes from them. First, we conduct similarity measurements between candidates $\mathbf S$ and memory prototypes $\mathbf C$. Then we determine the known cluster $\mathbf{S}'$ by:
\begin{equation}
\begin{split}
\mathbf{S'} &= {\mathrm{Match}}({{\mathbf C}, \mathbf{S}}) \\
&= \{{\mathbf s}_k|{\mathbf s}_k={\mathrm{argmax}}({\mathrm{sim}}({{\mathbf c}_k},{\mathbf S})), k \in [1,l]\},
\end{split}
\end{equation}
where ${\mathbf{S'}}=\{{\mathbf s}_1,{\mathbf s}_2,\cdots\, {\mathbf s}_l\}$ is a set of cluster centroids, each of which has a maximum similarity to one and only one non-overlapping class prototype ${{\mathbf c}_k}\in{\mathbf C}$. We consider ${\mathbf s}_k \in \mathbf{S \setminus S'}$ as the prototypes of novel classes and append them to Augmented-Memory. After the above prototype augmentation, the content of memory is extended to $\mathbf{C}=\{{\mathbf c}_1, {\mathbf c}_2, \cdots, {\mathbf c}_l, \cdots, {\mathbf c}_{l'} \}({l'} \ge l)$, and this operation can be considered an offline update of memory. 

\textbf{Center Contrastive loss.} 
After prototype matching and discovery, the same pseudo-labels are assigned to features from the same cluster according to the corresponding prototypes. At this point, the dataset is differentiated into (pseudo-)labeled features $\mathbf{F'}=\{\mathbf{f}'_1,\mathbf{f}'_2,\cdots \}$ and outlier features $\mathbf{E}=\{\mathbf{e}_1,\mathbf{e}_2,\cdots \}$. To further disentangle remaining outliers and enhance the intra-class compactness of (pseudo-)labeled features, we design a Center Contrastive Loss $\mathcal{L}_{Center}$ for instance-level contrastive learning with the help of the stored prototypes. 

Before each training iteration, we perform Global-Local Voting \cite{1} to select a reliable nearest neighbor in a mini-batch for each anchor and optimize the embedding space by enhancing the similarity between the anchor and its nearest neighbor. To alleviate the stochastic noise and avoid cluster dispersion, we do not crudely bring the nearest neighbor and the anchor closer together. Instead, we select different optimization strategies based on the anchor nature. In our method, if the anchor is (pseudo-)labeled, we force the nearest neighbor to move to the corresponding class prototype of the anchor; if the anchor is an outlier, we directly force the nearest neighbor to move to the anchor. Driven by the above setting, we design the following contrastive learning loss $\mathcal{L}_{Center}$:
\begin{equation}
\begin{split}
{\mathcal{L}_{Center}} =& {\mathcal{L}_{Center(a)}} + {\mathcal{L}_{Center(b)}} \\
=& - \frac{1}{|\mathbf{F'}|}\sum\limits_{\mathbf{f}'_a \in \mathbf{F'}} 
{\log({\|\mathbf{f}'_{nei}-{\mathrm{Pr}}(\mathbf{f}'_a)\|}_2)}\\  
& - \frac{1}{|\mathbf{E}|} \sum\limits_{\mathbf{e}_a \in \mathbf{E}}   
{\log({\|\mathbf{e}_{nei}-\mathbf{e}_a\|}_2)},
\end{split}
\end{equation} 
where $\mathrm{Pr}(\cdot)$ is the operation used to find the corresponding class prototype $\mathbf{c}_a$ of anchor $\mathbf{f}'_a$; $\mathbf{e}_{nei}$ and $\mathbf{f}'_{nei}$ denote the nearest neighbor of $\mathbf{e}_a$ and $\mathbf{f}'_a$, respectively; $|\cdot|$ denotes the number of features in set $\mathbf{F'}$ or $\mathbf{E}$ from a mini-batch. 

At the end of each training iteration, we also perform an online update for the Augmented-Memory. Specifically, all of the (pseudo-)labeled features $\mathbf{f'} \in \mathbf{F'}$ in a mini-batch are engaged in a momentum update \cite{40} to optimize their related class prototypes $\mathbf{c}$ at backward propagation stage, and the momentum coefficient $m$ is empirically set to 0.2:
\begin{equation}
\mathbf{c} \leftarrow m * \mathbf{c} + (1-m) * \mathbf{f'}.
\end{equation}
By alternately performing offline update or online update to the Augmented-Memory, we can effectively prevent overfitting and obtain comprehensive class prototypes to assist our instance-level disentanglement process.

\subsection{Training Strategy}
\begin{algorithm}[t]
    {\caption{An overview of our proposed DATA.}\label{alg1}
    \KwIn{$n$ labeled facial images ${X^l}=\{x_1^l,x_2^l, \cdots ,x_n^l\}$, \\
          $m$ unlabeled facial images ${X^u} = \{x_1^u,x_2^u, \cdots ,x_m^u\}$;}
    \textbf{Step~1: \ Before training}\;
    Memory Initialization~${\mathbf{C}} =\{ {{\mathbf c}_1,{\mathbf c}_2, \cdots ,{\mathbf c}_l} \}$;\\
    \For{$i$ \KwTo $MaxEpoch$}{
    \textbf{Step~2: \ Off-line Augmented-Memory update}\;
    Conduct DBSCAN cluster and augment prototypes\\ ${\mathbf{C}}=\{{\mathbf c}_1,{\mathbf c}_2, \cdots ,{\mathbf c}_l, \cdots ,{\mathbf c}_{l'} \}(l' \ge l)$;\\
        \For{$j$ \KwTo MaxIteration}{
        \textbf{Step~3: \ Feature extraction}\;
        $\mathbf{F}^l={\mathrm{E}}({X}^l), \mathbf{F}^u={\mathrm{E}}({X}^u)$; \\
        \textbf{Step~4: \ Feature-level disentanglement}\;
        $\mathbf{F}^g={\mathrm{MLP}}_g(\mathbf{F}), \mathbf{F}^s={\mathrm{MLP}}_s(\mathbf{F})$; \\
        \textbf{Step~5: \ Calculating deepfake bases}\;
        $\mathbf{O} = {\mathrm{GramSchmidt}}({\mathbf{F}^g})$;\\
        \textbf{Step~6: \ Instance-level disentanglement}\;
        $\mathbf{F'}=\{\mathbf{f}'_1,\mathbf{f}'_2,\cdots \},
         \mathbf{E}=\{\mathbf{e}_1,\mathbf{e}_2,\cdots \}$;\\
        \textbf{Step~7: \ Calculating loss and BP}\;
        $\mathcal{L} = \mathcal{L}_{CE}+{\eta _1}{\mathcal{L}_{Bases}} + 
        {\eta _2}{\mathcal{L}_{Center}} + \mathcal{R}$\;
        Gradient backpropagation;\\
        Online Augmented-Memory update.
        }}}
\end{algorithm}

Considering all the modules of DATA, we optimize the total network via the following loss function:
\begin{equation}
\mathcal{L} = {\mathcal{L}_{CE}} + {\eta _1}{\mathcal{L}_{Center}} + {\eta _2}{\mathcal{L}_{Bases}} +{\mathcal{R}},
\end{equation}
where $\mathcal{L}_{CE}$ and $\mathcal{R}$ are the cross-entropy loss and regularization terms included in the benchmark; $\eta _1$, $\eta _2$ are pre-defined hyperparameters used to adjust the weight of each item. Finally, to provide a clearer illustration, we summarize our entire workflow in Algorithm \ref{alg1}.

\section{EXPERIMENTS}
\subsection{Dataset and Evaluation Protocols}

\begin{table}[t]
\centering
\caption{The methods utilized in different protocols. StyleGAN2$^\ddag$ and StyleGAN2*: they belong to the same family of methods, but differ in the source of the real images used.}
\label{tab:data}
\scalebox{0.65}{
\begin{tabular}{llcccc}
\toprule
\textbf{Face Type} & \textbf{Method} & \textbf{Protocol-1} & \textbf{Protocol-2} & \textbf{Protocol-3} & \textbf{Protocol-4} \\
\midrule
\multirow{5}{*}{Identity Swap} 
    & Deepfakes~\tablefootnote{https://github.com/deepfakes/faceswap} & \checkmark & \checkmark & \checkmark & \checkmark \\
    & FaceSwap~\tablefootnote{https://github.com/MarekKowalski/FaceSwap/} & \checkmark & \checkmark & \checkmark & \checkmark \\
    & DeepFaceLab~\tablefootnote{https://github.com/iperov/DeepFaceLab} & \checkmark & \checkmark & \checkmark & \checkmark \\
    & FaceShifter~\cite{FaceShifter} & \checkmark & \checkmark & \checkmark & \checkmark \\
    & FSGAN~\cite{FSGAN} & \checkmark & \checkmark & \checkmark & \checkmark \\
\midrule
\multirow{5}{*}{Expression Transfer} 
    & Face2Face~\cite{Face2Face} & \checkmark & \checkmark & \checkmark & \checkmark \\
    & NeuralTextures~\tablefootnote{https://github.com/SSRSGJYD/NeuralTexture} & \checkmark & \checkmark & \checkmark & \checkmark \\
    & FOMM~\cite{FOMM} & \checkmark & \checkmark & \checkmark & \checkmark \\
    & ATVG-Net~\cite{ATVGNET} & \checkmark & \checkmark & \checkmark & \checkmark \\
    & Talking-Head-Video~\cite{Talking-Head-Video} & \checkmark & \checkmark & \checkmark & \checkmark \\
\midrule
\multirow{5}{*}{Attribute Manipulation} 
    & MaskGAN~\cite{MaskGAN} & \checkmark & \checkmark & \checkmark & \checkmark \\
    & StarGAN~\cite{StarGAN} & \checkmark & \checkmark & \checkmark & \checkmark \\
    & StarGAN2~\cite{StarGANv2} & \checkmark & \checkmark & \checkmark & \checkmark \\
    & SC-FEGAN~\cite{fegan} & \checkmark & \checkmark & \checkmark & \checkmark \\
    & FaceAPP~\tablefootnote{https://faceapp.com/app} & \checkmark & \checkmark & \checkmark & \checkmark \\
\midrule
\multirow{5}{*}{Entire Face Synthesis} 
    & StyleGAN~\cite{StyleGAN} & \checkmark & \checkmark & \checkmark & \checkmark \\
    & StyleGAN2$^\ddag$~\cite{StyleGAN2} & \checkmark & \checkmark & \checkmark & \checkmark \\
    & StyleGAN2*~\cite{StyleGAN2} & \checkmark & \checkmark & \checkmark & \checkmark \\
    & PGGAN~\cite{PGGAN} & \checkmark & \checkmark & \checkmark & \checkmark \\
    & CycleGAN~\cite{CycleGAN} & \checkmark & \checkmark & \checkmark & \checkmark \\
\midrule
\multirow{5}{*}{Diffusion-generated} 
    & LSGM~\cite{LSGM} &  &  & \checkmark & \checkmark \\
    & LatentDiffusion~\cite{LDM} &  &  & \checkmark & \checkmark \\
    & StableDiffusion~\cite{LDM} &  &  & \checkmark & \checkmark \\
    & MidJourney~\tablefootnote{https://www.midjourney.com/} &  &  & \checkmark & \checkmark \\
    & DALL-E~\tablefootnote{https://github.com/lucidrains/DALLE2-pytorch} &  &  & \checkmark & \checkmark \\
\midrule
\multirow{2}{*}{Real Face} 
    & Youtube-Real~\cite{ffpp} &  & \checkmark  &  & \checkmark \\
    & Celeb-Real~\cite{Celeb-df} &  & \checkmark &  & \checkmark \\
\bottomrule
\end{tabular}
}
\end{table}

\begin{figure}[t]
\centering
\includegraphics[width=0.45\textwidth]{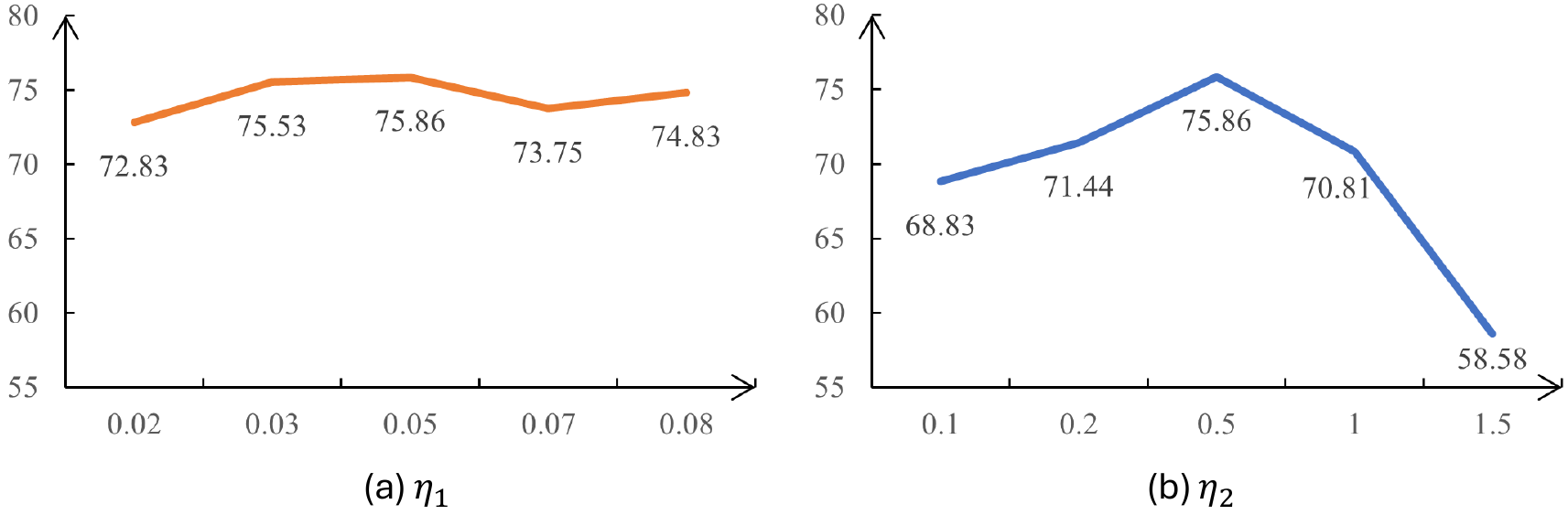}
\caption{The hyper-parameters analysis of loss function.}
\label{para}
\end{figure}

In this work, all of our experiments are conducted on two Open-World Semi-Supervised Deepfake Attribution datasets~\cite{1,65} along with different protocols. In summary, the OSS-DFA dataset consists of 2 real classes and 20 fake classes across 4 mainstream forgery types, and the OW-DFA++ dataset consists of 2 real classes and 25 fake classes across 5 mainstream forgery types. For more intuitive presentation, we summarized the generation techniques covered by different datasets in TABLE~\ref{tab:data}. It can be seen that Protocol-1/3 contains 20/25 fake classes and focuses solely on the performance of fake categories. Protocol-2/4 has been appended 2 real classes in addition to Protocol-1/3 and it focuses on the performance of the whole categories. Finally, following the benchmark, we divide the dataset into two parts with 80\% of the data allocated for training and 20\% for testing. Notably, this means the training set and the test set share the same label space. Furthermore, we validate the accuracy of our method DATA from different perspectives using three metrics: Accuracy (ACC), Normalized Mutual Information (NMI), and Adjusted Rand index (ARI) \cite{25-1,43}. 

\subsection{Implementation Details}
\begin{table*}[t]
\caption{The results of DATA and some state-of-the-art methods on OSS-DFA benchmark.}
\label{tab:comparison-1}
\resizebox{1.0\linewidth}{!}{
\begin{tabular}{l r cccccc c cccccc}
\toprule
\midrule
\multirow{3}{*}{Method}&\multirow{3}{*}{Venue}
&\multicolumn{6}{c}{Protocol-1}& &\multicolumn{6}{c}{Protocol-2}\\ 
\cmidrule{3-8} \cmidrule{10-15}
& &\multicolumn{3}{c}{Novel}&\multicolumn{3}{c}{All}    
& &\multicolumn{3}{c}{Novel}&\multicolumn{3}{c}{All}\\ 
\cmidrule{3-8} \cmidrule{10-15}
& &ACC&NMI&ARI&ACC&NMI&ARI& &ACC&NMI&ARI&ACC&NMI&ARI\\ 
\midrule
Lower Bound   & &{$40.96$}&{$46.43$}&{$24.05$}  &{$46.90$}&$63.18$&{$36.35$}& &{$46.48$}&{$48.44$}&{$31.49$} &$65.73$&$68.91$&$65.75$\\ 
Upper Bound   & &{$95.36$}&{$91.57$}&{$92.14$}  &{$96.68$}&$93.94$&{$93.59$}& &{$94.15$}&{$91.93$}&{$93.11$} &$96.83$&$93.80$&$95.05$\\ \midrule
DNA-Det\cite{7-2}       &AAAI'22&{$34.82$}&{$44.22$}&{$19.35$}  &{$34.99$}&$55.55$&{$24.89$}& &{$28.44$}&{$25.97$}&{$ 8.18$} &$54.37$&$50.10$&$31.45$\\
Openworld-GAN\cite{7-3} &CVPR'21&{$38.93$}&{$45.89$}&{$41.52$}  &{$57.62$}&$57.63$&{$47.47$}& &{$46.68$}&{$53.66$}&{$45.82$} &$69.26$&$58.60$&$61.09$\\
RankStats\cite{43}     &ICLR'20&{$49.94$}&{$56.05$}&{$39.76$}  &{$72.49$}&$73.63$&{$66.49$}& &{$45.26$}&{$52.44$}&{$30.17$} &$74.39$&$72.21$&$81.66$\\
ORCA\cite{25-1}          &ICLR'22&{$66.32$}&{$63.00$}&{$53.30$}  &{$80.81$}&$79.23$&{$74.05$}& &{$53.81$}&{$60.01$}&{$38.91$} &$78.99$&$78.04$&$83.80$\\
OpenLDN\cite{25-6}       &ECCV'22&{$45.83$}&{$51.05$}&{$38.12$}  &{$63.94$}&$71.38$&{$62.53$}& &{$42.23$}&{$50.66$}&{$28.86$} &$71.19$&$73.26$&$82.51$\\ 
NACH\cite{25-2}          &NIPS'22&{$70.13$}&{$67.10$}&{$56.63$}  &{$82.61$}&$81.98$&{$76.41$}& &{$53.92$}&{$58.49$}&{$38.73$} &$79.53$&$77.91$&$84.53$\\
CPL\cite{1}           &ICCV'23&{$71.89$}&{$68.20$}&{$59.37$}  &{$83.70$}&$82.31$&{$77.64$}& &{$59.92$}&{$63.90$}&{$43.75$} &$81.10$&$80.23$&$84.99$\\
MPSL\cite{65}           &IJCV'24&{$73.31$}&{$71.99$}&{$63.25$}  &{$84.31$}&$84.11$&{$78.97$}& &{$60.82$}&{$65.43$}&{$44.82$} &$82.17$&$81.92$&$86.61$\\ \midrule
\textbf{DATA (Ours)}    &\multicolumn{1}{c}{-}&{$\bm{75.86}$}&{$\bm{72.95}$}&{$\bm{64.90}$} &{$\bm{84.89}$}&$\bm{84.34}$&{$\bm{79.10}$}& &{$\bm{66.52}$}&{$\bm{71.98}$}&{$\bm{60.45}$} &$\bm{84.15}$&$\bm{84.19}$&$\bm{87.96}$\\ 
\midrule 
\bottomrule
\end{tabular}
}
\end{table*}
\begin{table*}[t]
\caption{The results of DATA and some state-of-the-art methods on OW-DFA++ benchmark.}
\label{tab:comparison-2}
\resizebox{1.0\linewidth}{!}{
\begin{tabular}{l r cccccc c cccccc}
\toprule
\midrule
\multirow{3}{*}{Method}&\multirow{3}{*}{Venue}&\multicolumn{6}{c}{Protocol-3}& &\multicolumn{6}{c}{Protocol-4}\\
\cmidrule{3-8} \cmidrule{10-15}
& &\multicolumn{3}{c}{Novel}&\multicolumn{3}{c}{All}& &\multicolumn{3}{c}{Novel}&\multicolumn{3}{c}{All}\\ 
\cmidrule{3-8} \cmidrule{10-15}
& &ACC&NMI&ARI&ACC&NMI&ARI& &ACC&NMI&ARI&ACC&NMI&ARI\\ 
\midrule
Lower Bound   
& &{$36.76$}&{$43.93$}&{$22.32$}&{$46.95$}&$62.13$&{$36.14$}& &{$36.11$}&{$40.43$}&{$19.47$} &$60.71$&$65.14$&$61.20$\\ 
Upper Bound   
& &{$96.69$}&{$94.45$}&{$94.29$}&{$97.70$}&$96.04$&{$95.64$}& &{$95.80$}&{$93.34$}&{$94.69$} &$97.92$&$96.18$&$98.23$\\ \midrule
DNA-Det\cite{7-2}&AAAI'22
&{$34.00$}&{$46.67$}&{$19.96$}  &{$46.93$}&$65.03$&{$34.20$}& &{$34.18$}&{$43.68$}&{$20.58$} &$58.98$&$66.26$&$55.62$\\
Openworld-GAN\cite{7-3}&CVPR'21
&{$41.69$}&{$42.04$}&{$38.31$}  &$55.98$&{$56.26$}&{$45.19$}& &{$41.76$}&{$48.22$}&{$35.37$} &$66.11$&$56.10$&$58.82$\\
RankStats\cite{43}&ICLR'20
&{$52.80$}&{$61.48$}&{$42.75$}  &{$74.72$}&$76.92$&{$68.17$}& &{$46.15$}&{$57.76$}&{$32.73$} &$77.65$&$77.74$&$88.59$\\
ORCA\cite{25-1}&ICLR'22
&{$55.36$}&{$60.06$}&{$46.56$}  &{$75.70$}&$78.71$&{$74.00$}& &{$48.49$}&{$55.77$}&{$36.14$} &$77.93$&$78.56$&$88.88$\\
OpenLDN\cite{25-6}&ECCV'22
&{$50.70$}&{$54.04$}&{$40.74$}  &{$71.62$}&$75.99$&{$67.60$}& &{$45.15$}&{$51.76$}&{$30.17$} &$71.42$&$74.16$&$85.56$\\ 
NACH\cite{25-2}&NIPS'22
&{$60.18$}&{$63.14$}&{$53.25$}  &{$77.97$}&$80.69$&{$76.45$}& &{$51.60$}&{$57.51$}&{$39.14$} &$79.48$&$80.24$&$90.12$\\
CPL\cite{1}&ICCV'23
&{$60.48$}&{$63.16$}&{$50.62$}  &{$78.37$}&$80.40$&{$76.13$}& &{$57.57$}&{$60.48$}&{$42.12$} &$81.78$&$81.10$&$89.60$\\
MPSL\cite{65}&IJCV'24
&{$62.79$}&{$68.65$}&{$54.98$}  &{$80.31$}&$83.58$&{$78.62$}& &{$59.84$}&{$66.38$}&{$50.71$} &$84.07$&$84.35$&$91.88$\\ \midrule
\textbf{DATA (Ours)}&\multicolumn{1}{c}{-}
&{$\bm{64.71}$}&{$\bm{70.62}$}&{$\bm{60.93}$} &{$\bm{82.00}$}&$\bm{84.79}$&{$\bm{79.18}$}& &{$\bm{61.90}$}&{$\bm{69.25}$}&{$\bm{57.42}$} &$\bm{84.11}$&$\bm{86.92}$&$\bm{92.86}$\\ 
\midrule 
\bottomrule
\end{tabular}
}
\end{table*}
Consistent with the OSS-DFA benchmark, all experiments are conducted on a single RTX3090 GPU. Simultaneously, we employ a pre-trained ResNet-50 \cite{41} as our feature extractor and utilize the dataset with a batch size of 128 and 50 epochs to train the entire network. Differently, we obtain the facial frames in a preprocessing step for less time consumption. During training, we use the Adam with a $2{e^{-4}}$ learning rate as our optimizer, and we decay the learning rate to 20\% of the original value every 10 epochs. In more detail, we set the hyperparameters of eps/min-samples in DBSCAN clustering to 0.03/6 and set the margin $\tau$ in Equation~(\ref{eq:loss_base}) to 20. As for hyperparameters used in the loss function, we first determine their approximate ranges based on experience and then conduct a more precise grid search to find their optimal settings. Considering the quality of pseudo labels will determine the upper bound of final accuracy, we give $\mathcal{L}_{Center}$ a larger weight than $\mathcal{L}_{Bases}$ in the total loss function. At the same time, due to the adversarial relationship between $\mathcal{R}$ and $\mathcal{L}_{CE}$, we give both of them the same weight $1$ following the benchmark. Based on the above analysis, we set the values of $\eta _1$ and $\eta _2$ to 0.5 and 0.05, respectively, and we present the detailed search process of them in Fig.\ref{para}. Also, it can be observed that $\eta _1$ influences the performance more significantly than $\eta _2$ which proves that the network itself focuses more on method-specific features while being minimally affected by method-general features.

\subsection{Comparison with Representative Methods}
\begin{table}[t]
\caption{ACC (\%) of cross-manipulation evaluation on the OSS-DFA dataset.}
\label{tab:comparison-3}
\centering
\resizebox{0.9\linewidth}{!}{
\begin{tabular}{l ccccc}
\toprule
\midrule
\multirow{2}{*}{Method}
&\multicolumn{2}{c}{Protocol-5}&&\multicolumn{2}{c}{Protocol-6}\\ 
\cmidrule{2-3}\cmidrule{5-6}
&Novel&Known&&Novel&Known\\ 
\midrule
% Backbone  &{40.00}&{64.54}&&{49.33}&{64.42} \\
MPSL      &{62.04}&{76.92}&&{59.68}&{71.06} \\
DATA (Ours)&$\bm{65.33}$&$\bm{79.30}$&&$\bm{63.00}$&$\bm{75.49}$\\
\midrule 
\bottomrule
\end{tabular}
}
\end{table}
\begin{table}[t]
\caption{ACC (\%) of robustness testing on the OSS-DFA dataset.}
\label{tab:robustness testing}
\centering
\resizebox{8.5cm}{!}{
\begin{tabular}{c ccccc}
\toprule
\midrule
\multirow{2}{*}{Method}&\multicolumn{4}{c}{ACC (\%)}\\ 
\cmidrule{2-5}
&Protocol-1&Protocol-2&Protocol-3&Protocol-4\\ 
\midrule
MPSL&{55.00}&{45.19}&{42.14}&{40.99}\\
DATA(Ours)&$\bm{58.00}$&$\bm{55.63}$&$\bm{43.91}$&$\bm{41.57}$\\
\midrule 
\bottomrule
\end{tabular}
}
\end{table}
To validate the effectiveness of our DATA, we compare it with various Open-World Semi-Supervised Learning~(OW-SSL) methods and open-world semi-supervised deepfake attribution~(OSS-DFA) method. Additionally, we also present the performances of the lower bound and upper bound, which signify the original performance of naive Resnet-50, and the fully supervised performance obtained by using labeled unknown samples for referring. The comparison results under different protocols are shown in Table~\ref{tab:comparison-1} and Table~\ref{tab:comparison-2}.
Obviously, the results of protocol-1/3 demonstrate that DATA outperforms the previous SOTA method MPSL by $2.55\%$/$1.92\%$ for novel classes and $0.58\%$/$1.69\%$ for all classes in terms of ACC. As for NMI and ARI, which measure the similarity of clustering results, DATA also demonstrates superior performance. At the same time, we are surprised to find that DATA still achieves favorable performance in protocol-2/4, whereas other methods generally perform lower. Here, DATA achieves a remarkable performance improvement of $5.7\%$/$2.06\%$ for novel classes and $1.98\%$/$0.06\%$ for all classes compared to MPSL. This phenomenon implicitly confirms that previous methods tend to extract forgery-irrelevant overfitting features to enhance the performance of the training set, making themselves difficult to handle a large number of real instances with fewer distribution biases. On the contrary, DATA is an expert in learning the true differences between various deepfake techniques and understanding the essence of `Deepfake.' In this way, our method exhibits good generalization on different categories, especially in the real category.

To validate the generalization ability, we also propose two new protocols, \textit{i.e.}, Protocol-5 and Protocol-6. These two protocols are built by removing an unknown forgery class from the training set of Protocol-1 and Protocol-2. That is, the test set contains classes that are entirely unseen during training. This `cross-manipulation' evaluation strategy introduces a significant challenge that has never been addressed in previous protocols. In Table~\ref{tab:comparison-3}, we compare our method with MPSL in this new setting. It can be seen that our method significantly outperforms the SoTA method and still achieves a 3\% improvement in ACC on the novel classes. Meanwhile, to validate the robustness of our method on low-quality data, we apply Gaussian blur to the test data and compare the accuracy of our method with MPSL under different protocols. The experimental results are presented in Table~\ref{tab:robustness testing}. Gaussian blur causes a significant performance drop, but our DATA framework still exhibits a smaller decline than the existing SoTA method MPSL. This indicates that our method demonstrates greater robustness when dealing with degraded images.

\begin{table}[t]
\caption{Ablation studies of DBE module in DATA.}
\label{tab:ablation}
\resizebox{8.5cm}{!}{
\begin{tabular}{cccc cccccc}
\toprule
\midrule
\multirow{2}{*}{Backbone}  
&\multirow{2}{*}{AMC}
&\multirow{2}{*}{\makecell{DBE\\({w/o}~$\mathcal{L}_{Base}$)}}  
&\multirow{2}{*}{\makecell{DBE\\({w/}~$\mathcal{L}_{Base}$)}} 
&\multicolumn{3}{c}{Novel}&\multicolumn{3}{c}{All}           \\ 
\cmidrule{5-10} 
&&&&ACC&NMI&ARI&ACC&NMI&ARI                                 \\ 
\midrule
\CheckmarkBold&&& 
&{$40.96$}&{$46.43$}&{$24.04$}&{$46.90$}&{$63.18$}&{$36.35$} \\ 
\CheckmarkBold&\CheckmarkBold&& 
&{$67.61$}&{$67.34$}&{$56.57$}&{$81.24$}&{$81.92$}&{$75.97$} \\ 
\CheckmarkBold&\CheckmarkBold&\CheckmarkBold&
&{$71.11$}&{$69.26$}&{$57.97$}&{$82.83$}&{$82.65$}&{$76.44$} \\
\CheckmarkBold&\CheckmarkBold&&\CheckmarkBold
&{$75.86$}&{$72.95$}&{$64.90$}&{$84.89$}&{$84.34$}&{$79.10$} \\ 
\midrule 
\bottomrule
\end{tabular}}
\end{table}
\begin{table}[t]
\caption{Ablation study of different loss functions in AMC.}
\label{tab:ablation-AMC}
\resizebox{8.5cm}{!}{
\begin{tabular}{ccc cccccc}
\toprule
\midrule
\multirow{2}{*}{Backbone+DBE}
&\multirow{2}{*}{\makecell{AMC\\({w/}~$\mathcal{L}_{Center(a)}$)}}
&\multirow{2}{*}{\makecell{AMC\\({w/}~$\mathcal{L}_{Center(b)}$)}}
&\multicolumn{3}{c}{Novel}&\multicolumn{3}{c}{All}           \\ 
\cmidrule{4-9} 
&&&ACC&NMI&ARI &ACC&NMI&ARI                                 \\ 
\midrule
\CheckmarkBold&&
&{$41.09$}&{$46.86$}&{$24.73$}&{$46.98$}&{$64.51$}&{$36.72$} \\ 
\CheckmarkBold&\CheckmarkBold&
&{$54.17$}&{$59.01$}&{$34.92$}&{$62.56$}&{$67.99$}&{$50.84$} \\ 
\CheckmarkBold&&\CheckmarkBold
&{$66.36$}&{$65.97$}&{$56.29$}&{$80.14$}&{$80.68$}&{$75.10$} \\ 
\CheckmarkBold&\CheckmarkBold&\CheckmarkBold
&{$75.86$}&{$72.95$}&{$64.90$}&{$84.89$}&{$84.34$}&{$79.10$} \\ 
\midrule 
\bottomrule
\end{tabular}}
\end{table}

\subsection{Ablation Study}

\begin{table}[t]
\caption{Comparison of Orthogonalization methods.}
\label{tab:orth}
\centering
\resizebox{8.5cm}{!}{
\begin{tabular}{c ccccc}
\toprule
\midrule
\multirow{2}{*}{Method}&\multicolumn{4}{c}{ACC (\%)}\\ 
\cmidrule{2-5}
&Protocol-1&Protocol-2&Protocol-3&Protocol-4\\ 
\midrule
SVD&{68.77}&{61.57}&{62.11}&{58.74}\\
PCA&{74.91}&{72.19}&{65.34}&{61.88}\\
Gram-Schmidt&$\bm{75.86}$&$\bm{66.52}$&$\bm{64.71}$&$\bm{61.90}$\\
\midrule 
\bottomrule
\end{tabular}
}
\end{table}
\begin{table}[t]
\caption{Comparison of pseudo-label generation methods.}
\label{tab:cluster}
\centering
\resizebox{8.5cm}{!}{
\begin{tabular}{c ccccc}
\toprule
\midrule
\multirow{2}{*}{Method}&\multicolumn{4}{c}{ACC (\%)}\\ 
\cmidrule{2-5}
&Protocol-1&Protocol-2&Protocol-3&Protocol-4\\ 
\midrule
K-means&{73.91}&{70.27}&{62.39}&{61.75}\\
DBSCAN&$\bm{74.88}$&$\bm{70.82}$&$\bm{65.54}$&$\bm{64.88}$\\
\midrule 
\bottomrule
\end{tabular}
}
\end{table}

As previously mentioned, we utilize the naive Resnet-50 as our backbone and enhance it through different modules of our method DATA. The whole network comprises the AMC module and the DBE module, each accompanied by an auxiliary loss. Specifically, the DBE module is designed to disentangle the forgery-related method-specific features and the Basis Contrastive loss $\mathcal{L}_{Bases}$ can further regularize the outputs of DBE. In the following AMC module, aided by the specialized Augmented-Memory, novel classes will be gradually discovered along with the whole training stage, and clustering pseudo labels will be produced with greater accuracy and less noise. Subsequently, all of the original labels and pseudo-labels are employed to construct the Center Contrastive Loss $\mathcal{L}_{Center}$ for facilitating the compactness within the same class. To explain the individual contributions clearly, we conduct the ablation experiments to validate the above 2 modules, and the corresponding experimental results are presented in Table~\ref{tab:ablation}.

\textbf{Ablation on AMC with $\mathcal{L}_{Center(a)}$ and $\mathcal{L}_{Center(b)}$.} As shown in the second row of Table \ref{tab:ablation}, $\mathrm {AMC}$ achieves a $26.65\%/34.34\%$ advantage in the accuracy of Novel/All classes over the baseline. To further verify the role of the loss $\mathcal{L}_{Center}$ included in AMC, we conducted ablation experiments on each component of the loss. As shown in Table~\ref{tab:ablation-AMC}, each component of the loss significantly contributes to the performance.

\begin{figure}
    \centering
    \includegraphics[width=0.46\textwidth]{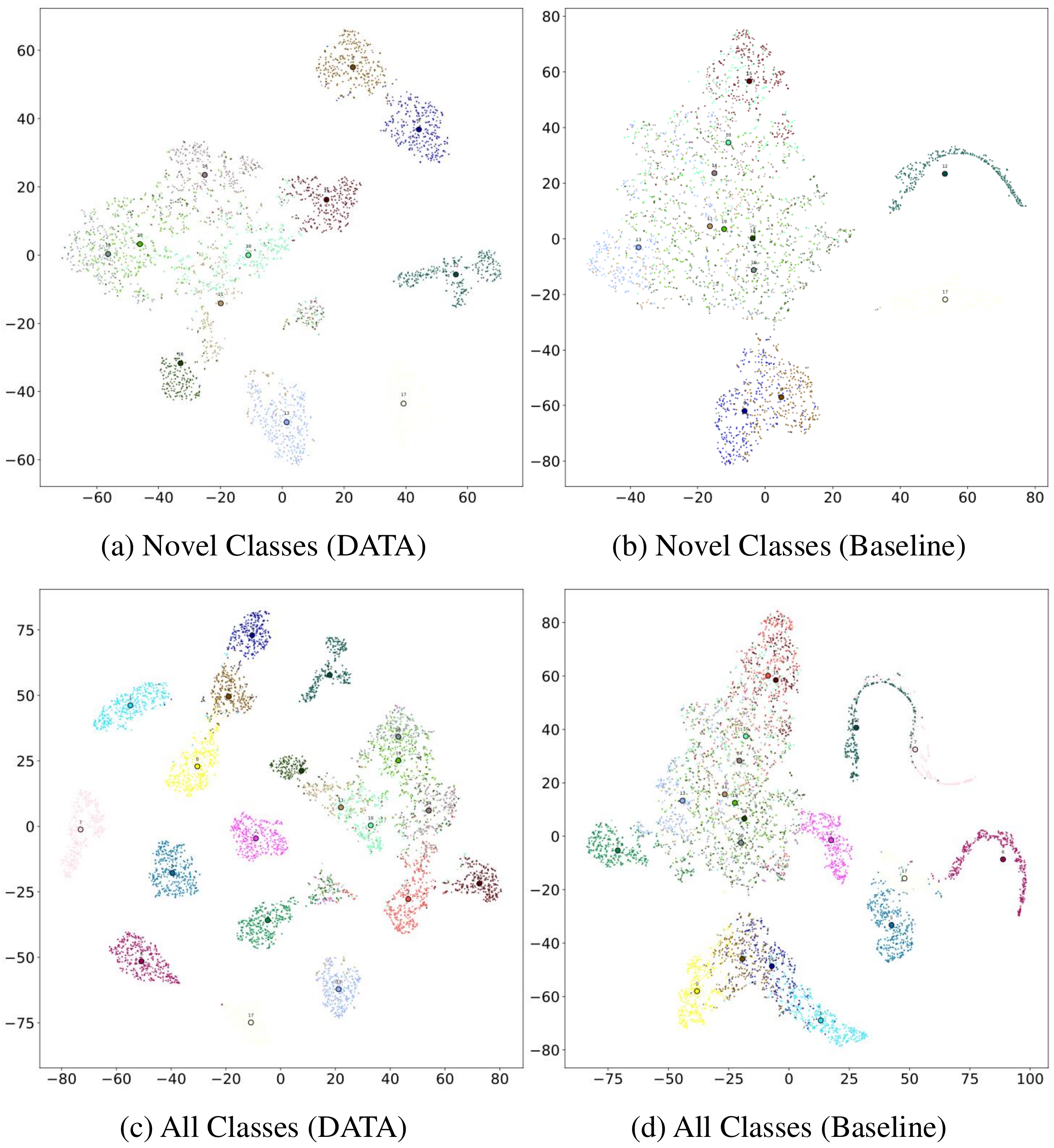}
    \caption{Comparison of t-sne maps.}
    \label{tsne}
\end{figure}

\begin{figure}[t]
    \centering
    \includegraphics[width=0.46\textwidth]{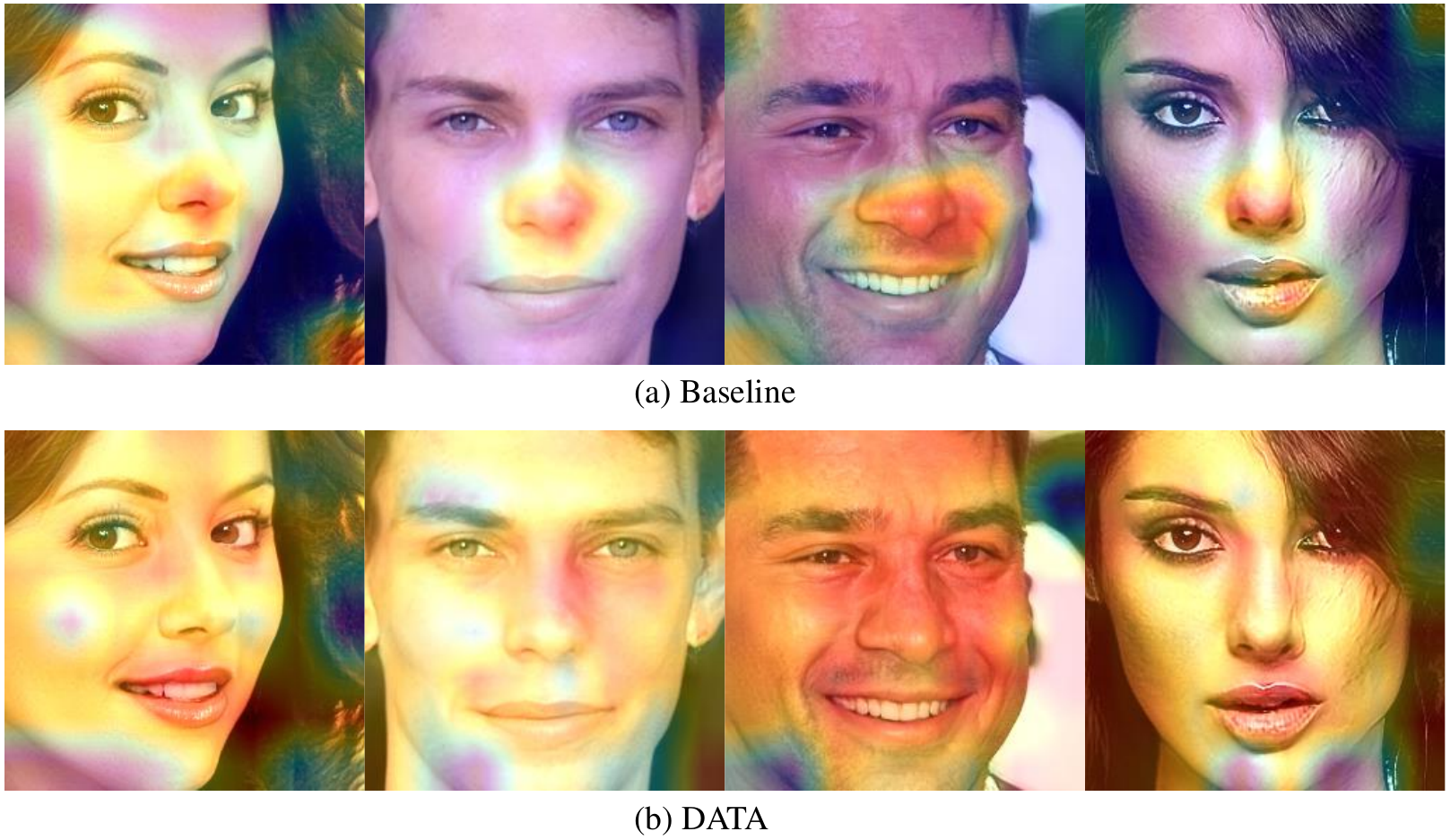}
    \caption{Comparison of CAM heat maps.}
    \label{fig:cam}
\end{figure}

\begin{figure}[t]
    \centering
    \includegraphics[width=0.46\textwidth]{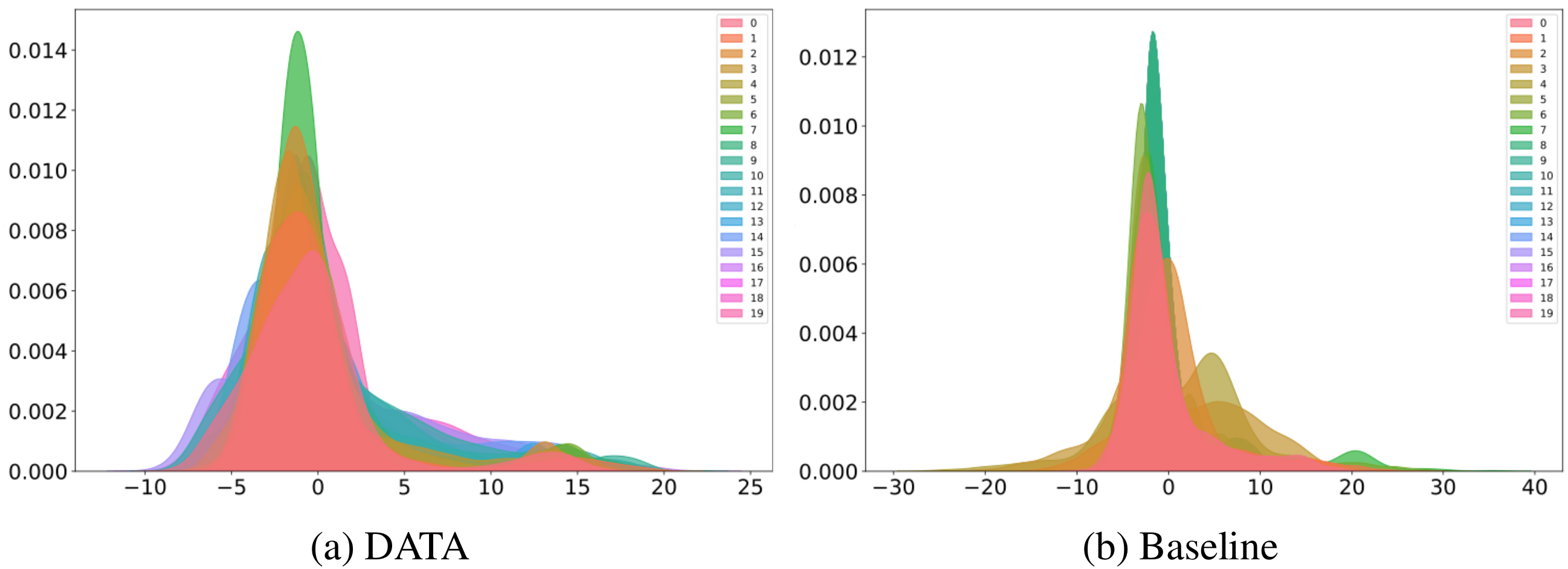}
    \caption{Comparison of feature distribution maps.}
    \label{distribution}
\end{figure}

\textbf{Ablation on DBE without $\mathcal{L}_{Bases}$.} Meanwhile, the $\mathrm {DBE}$ module further upgrades the performance of different metrics and improves the accuracy of Novel/All classes by $3.5\%/1.59\%$ based on $\mathrm {AMC}$. Although this improvement is not significant, we need to realize it is achieved by disentangling method-specific features without any constraint.

\textbf{Ablation on DBE with $\mathcal{L}_{Bases}$.} Finally, we represent the combination of $\mathrm {DBE}$ and $\mathcal{L}_{Bases}$ as $\mathrm{DBE^*}$. The addition of $\mathcal{L}_{Bases}$ still has a positive impact nearing the performance bottleneck. We find that $\mathcal{L}_{Bases}$ further boosts the final performance by $4.75\%/2.06\%$ based on AMC and DBE, showing its superiority in feature-level disentanglement.

\textbf{Ablation on different clustering or orthogonalization algorithms.} 
In order to construct a set of bases which can represent different forgery artifacts, we have experimented with different orthogonalization methods and selected Gram-Schmidt for its optimal performance. As shown in Table~\ref{tab:orth}, the performance of PCA and Gram-Schmidt is comparable, while SVD demonstrates certain limitations. One possible reason is that SVD retains all singular values, thereby introducing unnecessary noise.
To verify the effectiveness of the DBSCAN clustering algorithm, We employ `cluster accuracy' to measure the quality of pseudo-label. Specifically, cluster accuracy measures the alignment between the pseudo-labels generated by the clustering algorithm and the ground truth labels.
From Table~\ref{tab:cluster}, we can observe that DBSCAN achieves better performance. Therefore, we select it as the pseudo-label generation method for our DATA framework.

\subsection{Visualization}
To comprehensively showcase the superior performance of DATA, we compare our method with the baseline through t-sne maps, CAM heat maps, and feature distribution maps. In Fig.~\ref{tsne}, the first row depicts two t-SNE \cite{51} maps illustrating the distribution of novel classes, while the second row depicts two t-SNE maps illustrating the distribution of overall classes. A comparison between Fig.~\ref{tsne}(a) and Fig.~\ref{tsne}(b) reveals that our approach yields significantly clearer boundaries between the novel classes compared to the baseline. This observation suggests that the hierarchical contrastive learning we designed substantially improves the inter-class discrimination between novel classes. Furthermore, comparing Fig.~\ref{tsne}(c) and Fig.~\ref{tsne}(d), a notable enhancement in the intra-class compactness of known classes can be observed, consequently enhancing the classification accuracy of novel classes.

In Fig.~\ref{fig:cam}, we further illustrate all four types of deepfake technologies presented in the datasets: Attribute Manipulation, Entire Face Synthesis, Expression Transfer, and Identity Swap. In Fig.~\ref{fig:cam}(a), the first frame shows that the baseline primarily focuses on forgery-irrelevant areas although almost the entire face is influenced by attribute manipulation, indicating a significant overfitting issue; As for the second frame, due to the high quality of entire face synthesis techniques, baseline tends to lost in the hairlike features with lower generalization and overlooks the global coherence. Regarding the third frame, although some facial features related to expression transfer are highlighted, forgery-unrelated areas remain dominant. From the fourth frame, we can find that while identity transformation exposes numerous forgery clues, the baseline only relies on a small subset of information that cannot be considered as sufficient evidence. Conversely, Fig.~\ref{fig:cam}(b) shows DATA focuses on the global feature and is not affected by any background information. This means, by exploiting orthogonal deepfake basis, DATA pushes the network to deepen its understanding of `Deepfake' and leads it to utilize generalizable forgery-relevant features for multi-classification, rather than overfit to forgery-irrelevant patterns only applicable to the training dataset. Furthermore, by preserving early parameters, DATA avoids getting stuck in local optima and facilitates the extraction of global facial information.

In Fig.~\ref{distribution}, we visualize the distribution of output features. On the horizontal axis, a noticeable trend is observed: compared to the baseline, our method yields a more concentrated feature distribution. Here, only a small portion of the data deviates from the class prototype, with the deviation significantly smaller than that of the baseline. Additionally, on the vertical axis, it becomes apparent that our method leads to more data clustering around the class prototypes, indicating excellent intra-class compactness. Conversely, under the baseline, some class distributions appear inconsistent with the majority of other classes, suggesting that these classes did not utilize forgery-related features but instead relied on undesirable backdoor shortcuts. The observations above collectively imply that by integrating the use of DBE and AMC, DATA significantly promotes the forgery-related features exploration rather than overfitting to forgery-unrelated features or local features.

\subsection{Computational Requirements Analysis}

In Table~\ref{tab:computational}, we report the computational requirements (GFLOPs) and the number of parameters for our DATA and CPL. It can be observed that our approach does not require significantly more computational resources compared to other SoTA methods. For instance, in terms of GFLOPs, which measure inference speed, our method is nearly identical to CPL but our method achieves a 4\% improvement in ACC. At the same time, since the primary challenge in open-world semi-supervised scenarios is the lack of labeled samples, trading a slight increase in computational requirements for a reduction in labeling costs is also a feasible solution.
\begin{table}[t]
\centering
\caption{Computational requirements analysis.}
\label{tab:computational}
\resizebox{5cm}{!}{
\begin{tabular}{l cc}
\toprule
\midrule
Method &GFLOPs &Params\\
\midrule
CPL         &{$5.33$} &{$25.6$}M \\ 
% MPSL        &{$6.49$} &{$28.2M$} \\ 
DATA (Ours) &{$5.34$} &{$31.0$}M \\ 
\midrule 
\bottomrule
\end{tabular}}
\end{table}

\section{Conclusion}
The proposed DATA framework addresses both overfitting issues, led by dataset biases, and annotation pressures, from the rapidly emerging novel techniques. Additionally, through the collaboration of the Deepfake Basis Exploration module and the Augmented-Memory based Clustering module, the final classification results rely more on well-generalized forgery-relevant features, which are beneficial for identifying novel classes in open-world scenarios. In the future, we will conduct a more detailed review of the deepfake generation methods to seek theoretical support for the construction of forgery bases and uncover more effective utilization methods. Meanwhile, the newly added experimental results indicate that the existing semi-supervised methods still struggle to handle the cross-manipulation scenario. Therefore, we plan to conduct related research in this potential area.

\bibliographystyle{IEEEtran}
\bibliography{DFA-TMM}

\begin{IEEEbiography}[{\includegraphics[width=1in,height=1.25in,clip,keepaspectratio]{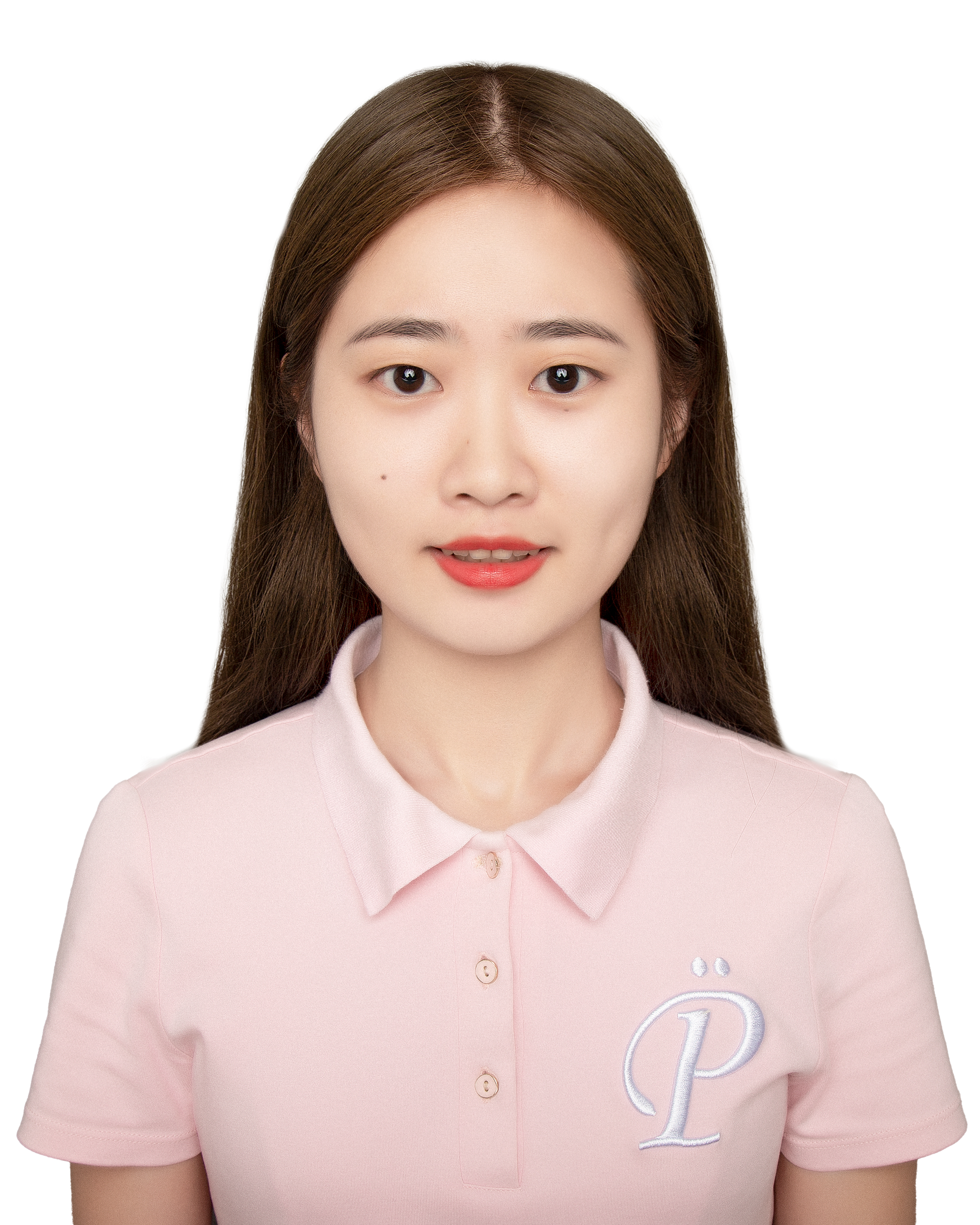}}]{Ming-Hui Liu} is currently pursuing the Ph.D. degree in software engineering at the School of Software, Shandong University, Jinan, China. Her research interests include computer vision and she focuses on the subfields of deepfake detection and person re-identification. She serves as a reviewer for various international conferences and journals, e.g. MM, IEEE IoT, and IEEE TETCI.
\end{IEEEbiography}
% \vspace{-2em}
\begin{IEEEbiography}[{\includegraphics[width=1in,height=1.25in,clip,keepaspectratio]{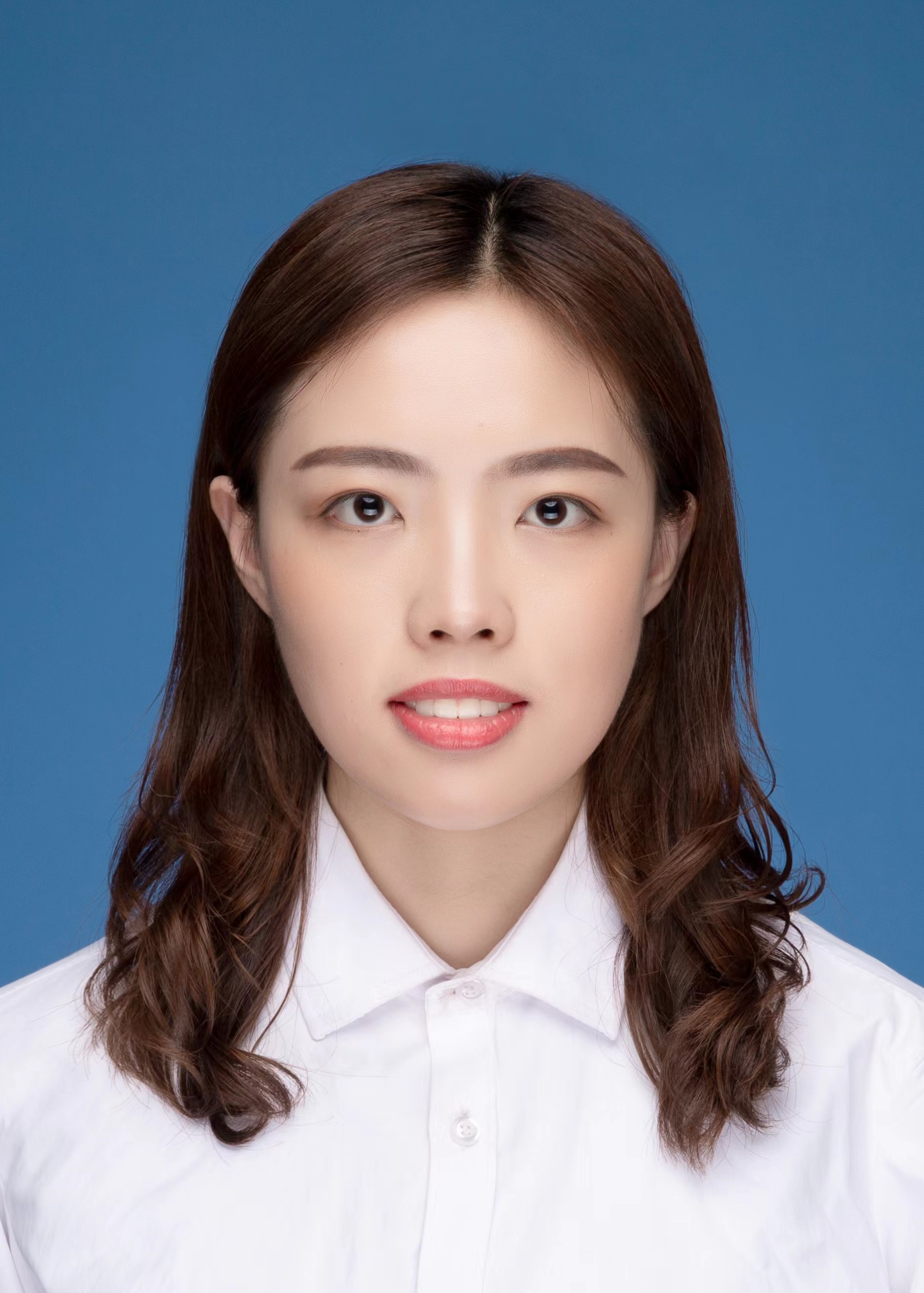}}]{Xiao-Qian Liu} received the M.S. degree in Control engineering in 2020 from Shandong University, China. She is currently pursuing the Ph.D. degree in artificial intelligence at the School of Software, Shandong University, Jinan, China. Her research interests include deep learning, computer vision, domain adaptation, and OCR.
\end{IEEEbiography}
\begin{IEEEbiography}[{\includegraphics[width=1in,height=1.25in,clip,keepaspectratio]{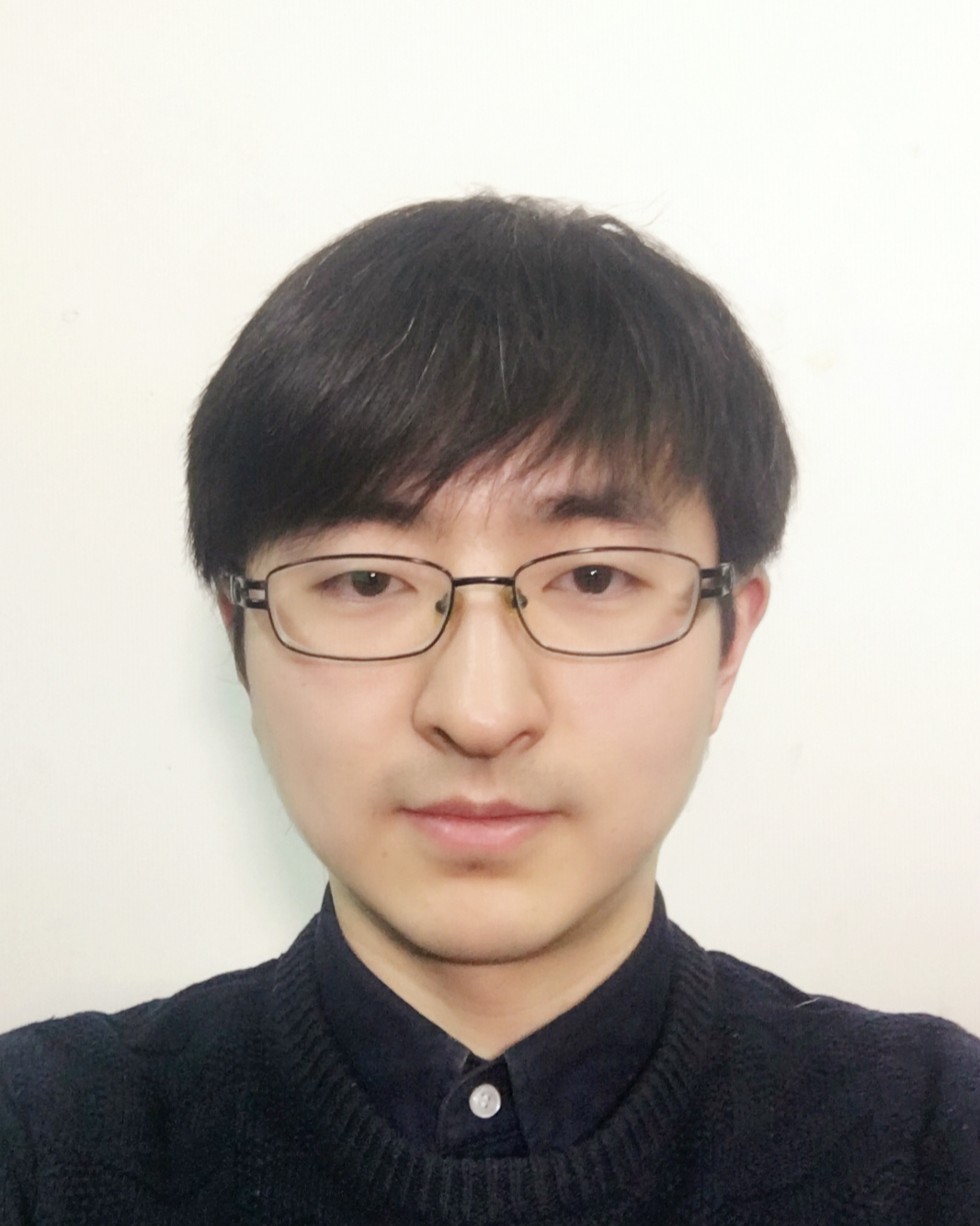}}]{Xin Luo} received the Ph.D. degree in computer science from Shandong University, Jinan, China, in 2019. He is currently an assistant professor with the School of Software, Shandong University, Jinan, China. His research interests mainly include machine learning, multimedia retrieval and computer vision. He has published over 20 papers on TIP, TKDE, ACM MM, SIGIR, WWW, IJCAI, et al. He serves as a Reviewer for ACM International Conference on Multimedia, International Joint Conference on Artificial Intelligence, AAAI Conference on Artificial Intelligence, the IEEE Transactions on Cybernetics, Pattern Recognition, and other prestigious conferences and journals.
\end{IEEEbiography}
\begin{IEEEbiography}[{\includegraphics[width=1in,height=1.25in,clip,keepaspectratio]{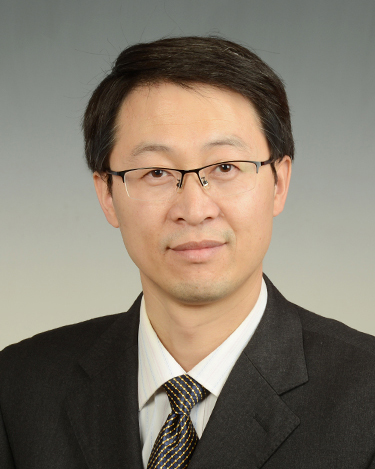}}]{Xin-Shun Xu} is currently a professor with the School of Software, Shandong University. He received his M.S. and Ph.D. degrees in computer science from Shandong University, China, in 2002, and Toyama University, Japan, in 2005, respectively. He joined the School of Computer Science and Technology at Shandong University as an associate professor in 2005, and joined the LAMDA group of Nanjing University, China, as a postdoctoral fellow in 2009. From 2010 to 2017, he was a professor at the School of Computer Science and Technology, Shandong University. He is the founder and the leader of MIMA (Machine Intelligence and Media Analysis) Lab of Shandong University. His research interests include machine learning, information retrieval, data mining and image/video analysis and retrieval. He has published in TIP, TKDE, TMM, TCSVT, AAAI, CIKM, IJCAI, MM, SIGIR, WWW and other venues. He also serves as an SPC/PC member or a reviewer for various international conferences and journals, e.g. AAAI, CIKM, CVPR, ICCV, IJCAI, MM, TCSVT, TIP, TKDE, TMM and TPAMI.
\end{IEEEbiography}

\vspace{11pt}
\vfill
\end{document}